\documentclass{article} 
\usepackage{iclr2027_conference,times}

\newif\ifarxiv

\arxivtrue          

\ifarxiv
    \iclrfinalcopy
\fi

\usepackage{amsmath,amsfonts,bm}

\def\eqref#1{equation~\ref{#1}}

\def\1{\bm{1}}

\DeclareMathAlphabet{\mathsfit}{\encodingdefault}{\sfdefault}{m}{sl}
\SetMathAlphabet{\mathsfit}{bold}{\encodingdefault}{\sfdefault}{bx}{n}

\usepackage{multirow}
\usepackage{hyperref}
\usepackage{url}
\usepackage{graphicx}
\graphicspath{{figures/}}
\usepackage{booktabs}

\title{From Reweighting to Rewriting: Unlocking the Intervention Effects of Influential Samples in Training Data Attribution}

\author{
\makebox[0.96\textwidth][c]{
Yuzhang Luo\textsuperscript{1,*} \quad
Chenpeng Wang\textsuperscript{2,*} \quad
Jianhui Chen\textsuperscript{1,*} \quad
Liangming Pan\textsuperscript{1,3,\textdagger}
}
\\
\makebox[0.96\textwidth][c]{
\textsuperscript{1}State Key Laboratory of Multimedia Information Processing, Peking University
}
\\
\makebox[0.96\textwidth][c]{
\textsuperscript{2}YiXin-AILab, YIXIN, Beijing, China
}
\\
\makebox[0.96\textwidth][c]{
\textsuperscript{3}Beijing Academy of Artificial Intelligence, Beijing, China
}
\\
\makebox[0.96\textwidth][c]{
\texttt{luoyuzzhang@stu.pku.edu.cn}
\quad
\texttt{liangmingpan@pku.edu.cn}
}
\\
\makebox[0.96\textwidth][c]{
\texttt{wangchenpeng@yxqiche.com}
\quad
\texttt{jianhuichennlp@gmail.com}
}
}

\begin{document}

\maketitle

\ifarxiv
    \fancyhead{}
    \renewcommand{\headrulewidth}{0pt}
\fi

\ifarxiv
\begingroup
\renewcommand{\thefootnote}{\fnsymbol{footnote}}
\footnotetext[1]{Equal contribution. \qquad
\textsuperscript{\textdagger}Corresponding author.}
\endgroup
\fi

\begin{abstract}

Training data attribution (TDA) aims to identify training examples that shape model behavior, but its intervention value depends on both which examples are selected and how they are modified. Influence functions (IF) estimate behavioral changes under infinitesimal reweighting, yet IF-selected examples often show limited advantages over random selection under conventional weight-based interventions. This raises the question of whether influential examples lack intervention value or whether reweighting fails to realize their behavioral leverage. We introduce influence-guided response rewriting, which uses IF to identify intervention targets and replaces their responses with behavior-aligned or behavior-opposed supervision while keeping instructions fixed. Across four open-weight LLMs, we compare rewriting and reweighting on the same influence-selected examples using epistemic abstention as our primary testbed. Response rewriting produces stronger, more persistent, and bidirectional behavioral shifts, while reweighting the same examples yields weak and inconsistent effects. Further analyses show that influence-selected examples provide greater rewriting leverage than alternative selectors, with changes remaining concentrated on target-relevant behaviors. The same qualitative contrast extends to safety refusal. These results distinguish the local reweighting effects captured by influence estimates from the broader intervention leverage of the examples they identify, motivating intervention-aware evaluation of TDA methods.
\end{abstract}

\section{Introduction}
\label{sec:introduction}

Large language models (LLMs) acquire diverse capabilities and behaviors from their training data, yet which training examples give rise to these behaviors remains poorly understood. Training data attribution (TDA) addresses this question by assigning training examples scores that quantify their contribution to a target model quantity
\citep{koh2017understanding,pruthi2020estimating,guo2021fastif}.
Influence functions (IF) provide one such approach by estimating how that quantity would change if a training example were infinitesimally reweighted, and have increasingly been applied to attribute LLM predictions, capabilities,
and behaviors
\citep{grosse2023studyinglargelanguagemodel,deng:hal-05230469}.

Beyond retrospective attribution, a central practical goal of TDA is to support actionable data interventions. A common approach is to use IF to identify training examples that strongly affect a target behavior, and then upweight or remove those examples in training to steer the model. However, recent studies find that IF-selected examples often provide little advantage over random selection under such weight-based interventions in modern LLMs~\citep{li2025influence,lee2026datafiltering}. Yet this negative result conflates two distinct questions: whether IF identifies the right examples to intervene on, and whether the intervention itself can effectively alter what those examples teach the model. It therefore remains unclear whether IF-selected examples are intrinsically poor intervention targets, or whether weight-based interventions simply fail to realize their behavioral leverage.

\begin{figure}[t]
\centering
\includegraphics[width=0.95\linewidth]{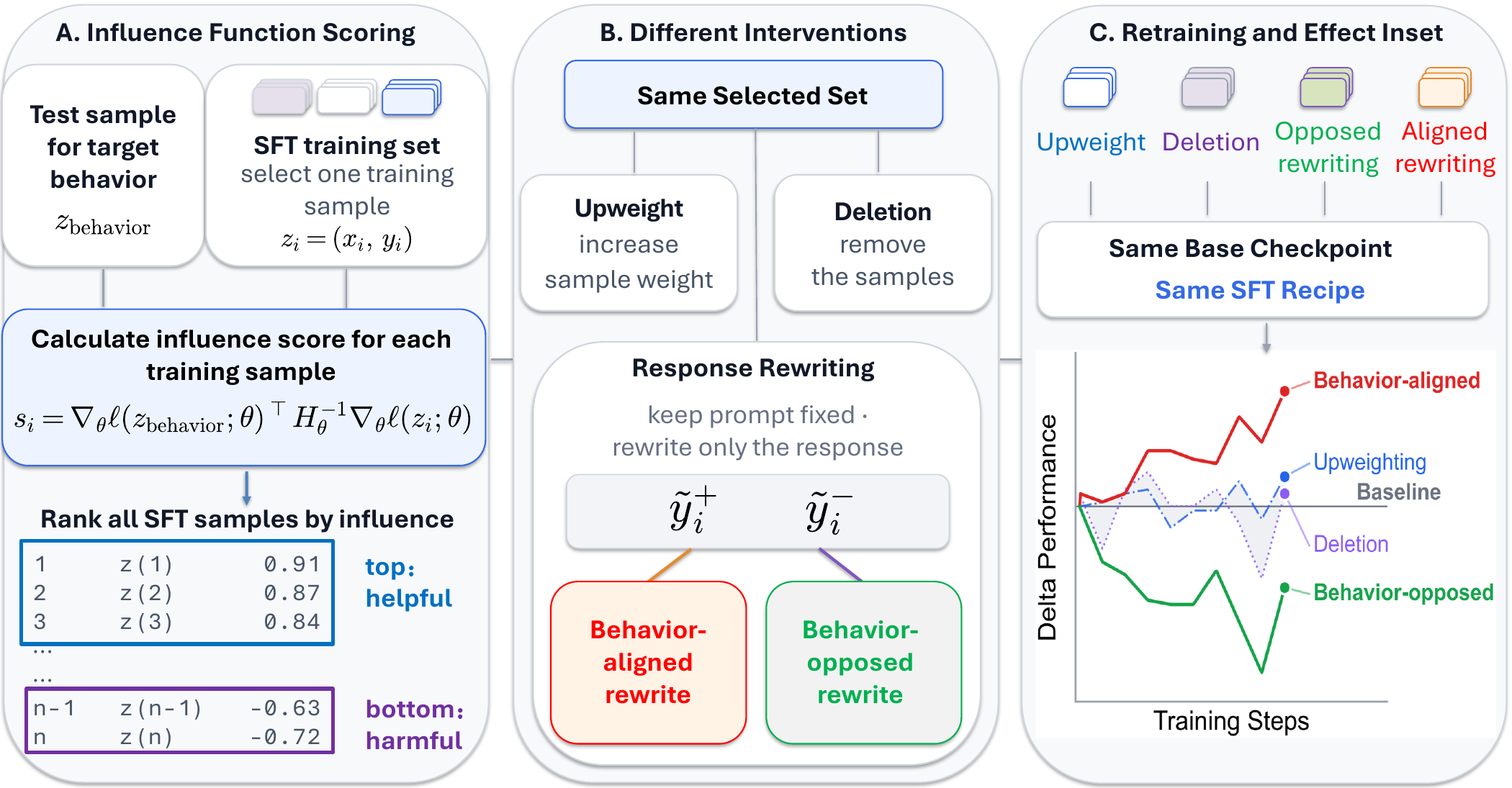}
\caption{
Overview of our framework. IF identifies training examples
associated with a target behavior, which we intervene on through weight-based
operations or response rewriting.
}
\label{fig:overview}
\end{figure}

We study this problem in the supervised fine-tuning (SFT) setting, where intervention need not be limited to changing how much an example contributes during training. Instead, we can directly modify what the example teaches the model by rewriting its response while keeping the instruction fixed. Motivated by this distinction, we propose \emph{influence-guided response rewriting}, where IF determines \emph{where} to intervene and response rewriting determines \emph{what} behavioral signal to provide~(Figure~\ref{fig:overview}). By rewriting responses to either encourage or discourage a target behavior, this framework both provides an actionable form of TDA and allows us to test whether influence-selected examples contain intervention leverage that is obscured by conventional weight-based interventions.

To investigate whether influence-guided response rewriting can reveal such hidden intervention leverage, we compare it with conventional weight-based interventions and matched controls. This comparison allows us to disentangle the value of influence-guided example selection from the limitations of a particular intervention strategy. More broadly, our study reframes TDA intervention from asking whether influential examples respond to reweighting, to asking whether influence scores identify training examples with actionable behavioral potential and how such potential can be effectively realized. To be specific, we make the following three contributions:






\begin{itemize}

\item \textbf{Influence-guided response rewriting.}
We introduce a supervision-level intervention framework for SFT that decouples example selection from intervention design. Specifically, influence functions identify high-leverage training examples, while response rewriting modifies the behavioral signal provided by these examples.

\item \textbf{Revealing hidden behavioral leverage of influential examples.}
We evaluate influence-guided interventions across four open-weight LLMs, primarily on language-model abstention \citep{zhang2024r,wen2025know,kirichenko2026abstentionbench}. Compared with conventional reweighting and matched controls, response rewriting consistently achieves stronger, more stable, and bidirectional behavioral shifts across model families and training stages. These results show that influential examples can contain substantial behavioral leverage that is not effectively realized through weight-based interventions. We further observe similar trends for safety-related refusal.

\item \textbf{Characterizing the source and scope of intervention leverage.}
Through controlled analyses, we investigate why influence-guided rewriting is effective. We show that response rewriting redirects the local supervision signal associated with influential examples, while maintaining target specificity and avoiding degradation of model capabilities.

\end{itemize}

\section{Related Work}
\label{sec:related_work}

\subsection{Training Data Attribution}

Training data attribution aims to quantify the influence of specific training examples on model behavior, with influence functions (IF) providing a foundational approach in neural networks \citep{koh2017understanding}. Scalable curvature approximations such as EK-FAC \citep{george2018fast} have enabled IFs to trace the training origins of LLM behaviors and capabilities
\citep{grosse2023studyinglargelanguagemodel,kou2025which}.
Influence-based attribution is commonly evaluated or applied through data reweighting and filtering \citep{chen2026mechanistic,Kowal2026ConceptIL,lee2026influence}, yet recent studies have found that influence estimates can correspond weakly to the effects of these interventions in LLMs \citep{bae2022if,li2025influence,lee2026datafiltering}.
Prior work has also explored modifying influential training examples directly, for example through influence-guided relabeling in classification \citep{kong2022resolving,banerjee2024inffeed}. Most closely related to our work, \emph{Infusion} uses influence functions to select and perturb existing training examples for targeted data-poisoning attacks \citep{DBLP:journals/corr/abs-2602-09987}. However, its reliable behavior changes are confined to vision settings and perturbation effects further diminish under longer training on transformers and language models. In contrast, we study semantic response rewriting in realistic LLM supervised fine-tuning, systematically compare deletion, upweighting, and rewriting on the same influence-selected examples, and track how their behavioral effects evolve throughout the training trajectory.

\subsection{Language Model Abstention}
Abstention refers to a model's ability to refrain from providing a definitive answer when a query cannot be reliably resolved. 
Recent large-scale evaluations show that the abstention capabilities of modern LMs remain poor across diverse forms of unanswerability  \citep{wen2025know,kirichenko2026abstentionbench}. Existing approaches address this problem from several directions, including uncertainty estimation, calibration, probing internal representations, prompting, and abstention-aware post-training
\citep{DBLP:journals/corr/abs-2207-05221,Tomani2024UncertaintyBasedAI,lavi2026detecting}.
In particular, refusal-aware instruction-tuning can shape abstention through constructing or replacing responses with abstention-aware targets \citep{yang2024alignment,zhang2024r}. More recent work improves refusal-aware tuning through knowledge-aware data modification and training-dynamics analysis
\citep{DBLP:conf/aaai/ZhuMWGWLH25}, or gradient-based sample selection and adaptive weighting
\citep{zhu2025grait}. However, prior work points out that instruction-tuning struggles to generalize abstention capability across domains and model settings
\citep{feng2024don} and fine-tuning can even erode abstention capabilities
\citep{kirichenko2026abstentionbench}. These findings leave the data-level mechanisms through which abstention evolves during realistic SFT underexplored.
We study this problem from a training data attribution perspective, tracing abstention behavior to individual SFT examples and systematically comparing deletion, upweighting, and response rewriting as alternative interventions.

\section{Methodology}
\label{sec:method}

Our goal is to determine whether influence-selected SFT examples possess behavioral intervention leverage beyond what is revealed by conventional weight-based interventions. To isolate these two factors, we keep the attribution rule fixed and vary how the selected examples are intervened on. Specifically, IFs determine which training examples are selected, while the intervention either changes the strength of their original supervision or rewrites the supervision they provide.

\subsection{Influence Functions for SFT}
\label{sec:influence}
Let $\mathcal{D}_{\mathrm{train}}=\{z_i=(x_i,y_i)\}_{i=1}^{N}$ denote an
SFT dataset, where $x_i$ is an instruction and $y_i$ its response. We use the
standard response-only loss
$\mathcal{L}(z_i,\theta)
=-\sum_t\log p_\theta(y_{i,t}\mid x_i,y_{i,<t})$.
Influence functions estimate how infinitesimally changing the training weight
of an example affects the learned model parameters
\citep{koh2017understanding}. For a training example $z_i$, consider
\begin{align}
\theta(\epsilon)
&=
\arg\min_{\theta}
\left[
\frac{1}{N}\sum_{j=1}^{N}\mathcal{L}(z_j,\theta)
+
\epsilon\mathcal{L}(z_i,\theta)
\right],
\qquad
\left.
\frac{d\theta(\epsilon)}{d\epsilon}
\right|_{\epsilon=0}
=
-H_{\theta}^{-1}
\nabla_{\theta}\mathcal{L}(z_i,\theta),
\label{eq:weight_perturb}
\\[-2pt]
\mathcal{I}_{f}(z_i)
&=
\left.
\frac{d f(\theta(\epsilon))}{d\epsilon}
\right|_{\epsilon=0}
=
-
\nabla_{\theta}f(\theta)^{\top}
H_{\theta}^{-1}
\nabla_{\theta}\mathcal{L}(z_i,\theta),
\qquad
\text{for any differentiable } f(\theta).
\label{eq:influence}
\end{align}
Here $H_{\theta}$ denotes the Hessian of the SFT objective, with all quantities
evaluated at the unperturbed solution $\theta=\theta(0)$. Under this convention,
a positive influence score $\mathcal{I}_{f}(z_i)>0$ predicts that locally
upweighting $z_i$ increases $f(\theta)$, while downweighting it decreases
$f(\theta)$. Conversely, a negative score $\mathcal{I}_{f}(z_i)<0$ predicts
that upweighting $z_i$ decreases $f(\theta)$, while downweighting it increases
$f(\theta)$.

Following prior LLM-scale influence-function work
\citep{george2018fast,grosse2023studyinglargelanguagemodel,kou2025which}, 
we use EK-FAC to approximate the required inverse-curvature computation.
Derivation and implementation details are provided in
Appendix~\ref{app:influence}.

\subsection{Influence-Guided Data Interventions}
\label{sec:interventions}

\subsubsection{Behavior Attribution}
\label{sec:behavior}
Given a query set
$\mathcal{D}_{\mathrm{tar}}=\{(x_j^{q},y_j^{q})\}_{j=1}^{M}$
representing a target behavior, we define its mean response log-likelihood,
$f(\theta)=\frac{1}{M}\sum_{j=1}^{M}\frac{1}{T_j}
\sum_{t=1}^{T_j}\log p_{\theta}(y_{j,t}^{q}\mid x_j^{q},y_{j,<t}^{q})$,
as a differentiable behavioral proxy.
We rank the SFT examples by $\mathcal{I}_{f}(z_i)$ and define
$\mathcal{S}^{\mathrm{helpful}}_k=\operatorname{TopK}_i\mathcal{I}_{f}(z_i)$
and
$\mathcal{S}^{\mathrm{harmful}}_k=\operatorname{BottomK}_i\mathcal{I}_{f}(z_i)$.
We refer to these sets as \emph{supposedly helpful} and
\emph{supposedly harmful}, respectively, because the labels describe only the
local effects predicted for their original supervision under infinitesimal
reweighting.

\subsubsection{Intervention Design}

We apply two families of interventions to the same influence-selected examples. The first changes the strength of the original supervision and directly follows the local reweighting interpretation of influence functions. The second changes
the content of the supervision by keeping the selected instruction fixed while rewriting its response. Comparing the two allows us to test whether the usefulness of influence-selected examples is limited to reweighting original supervision, or whether these examples exhibit broader behavioral leverage under changes to the supervision they provide.

\paragraph{Reweighting original supervision.}
For a selected set $\mathcal{S}$, we optimize the weighted SFT objective $\mathcal{L}_{\alpha}(\theta)=\sum_i w_i(\alpha)\ell_i(\theta)$, where $w_i(\alpha)=\alpha$ if $i\in\mathcal{S}$ and $1$ otherwise. Here, $\alpha$ controls the relative contribution of selected examples: $\alpha>1$ corresponds to \textbf{upweighting}, while $\alpha=0$ corresponds to \textbf{deletion}. These interventions preserve the original
responses of the selected examples and modify only how strongly their existing
supervision contributes during training.

\paragraph{Rewriting supervision.}
We next consider interventions that directly change what a selected example
teaches the model. For each selected example $z_i=(x_i,y_i)$, we keep its
instruction $x_i$ fixed and replace its response according to
$\mathcal{R}_{d}(x_i,y_i)=(x_i,\widetilde{y}^{\,d}_i)$, where $d\in\{\mathrm{align},\mathrm{opp}\}$ denotes supervision that encourages the target behavior or its opposite. We refer to this intervention as
\emph{influence-guided response rewriting}.

Unlike reweighting, response rewriting changes the gradient contributed by the
selected example and therefore should not be interpreted as a finite
realization of the local influence prediction in Eq.~\ref{eq:influence}.
Instead, influence is used only to determine which examples to modify. This
distinction is central to our study: if rewriting influence-selected examples
produces larger behavioral changes than applying the same rewriting procedure
to matched random examples, then the influence ranking identifies examples
with intervention leverage that extends beyond reweighting their original
supervision.

\subsubsection{Evaluation Protocol}

For an intervention $\mathcal{O}$ applied to a selected set $\mathcal{S}$, let $\Delta(\mathcal{O},\mathcal{S})
=
B(\theta_{\mathcal{O},\mathcal{S}})
-
B(\theta_{\mathrm{base}})
$
denote the resulting change in a behavioral metric $B$. We evaluate each intervention along two complementary dimensions.

\paragraph{Directional effectiveness.}
We first ask whether the intervention moves the target behavior in its intended
direction. Under the local influence prediction, upweighting supposedly helpful
examples or deleting supposedly harmful examples should strengthen the target
behavior, while the reverse operations should weaken it. For response
rewriting, aligned and opposed responses should induce behavioral shifts in the
corresponding directions.

\paragraph{Selection advantage.}
We then compare each intervention on influence-selected examples with the same
intervention applied to matched random examples. This tests whether
influence-guided selection identifies examples with greater behavioral leverage
than arbitrary training examples under the same intervention.

We track both directional effectiveness and selection advantage throughout SFT,
allowing us to distinguish persistent intervention effects from effects that
arise only at isolated training checkpoints.
\section{Experiments}
\label{sec:abstention}

We instantiate our framework on epistemic abstention, a behavior for which the model should refrain from answering when a query cannot be reliably resolved.

\subsection{Experimental Setup}
\label{sec:abstention_setup}

\paragraph{Abstention target and evaluation.}
Following the scenario taxonomy of AbstentionBench~\citep{kirichenko2026abstentionbench},
we construct the target function $f(\theta)$ using 300 held-out abstention queries drawn primarily from two scenarios: \emph{answer unknown}, where no documented or commonly agreed-upon answer exists, and \emph{false premise}, where the query is predicated on a false statement. These target queries are disjoint from both the SFT data and evaluation sets. Unless otherwise specified, we use \emph{answer unknown} as the primary evaluation scenario for our training-dynamics analysis. Detailed query construction and other
scenario-wise results are provided in Appendix~\ref{app:rewrite} and Appendix~\ref{app:targetspecific}.

We mainly evaluate the resulting models using \emph{abstention recall}, defined as the fraction of unanswerable evaluation queries on which the model abstains:
\begin{equation}
\mathrm{Recall}
=
\frac{
\#\{\text{unanswerable queries on which the model abstains}\}
}{
\#\{\text{unanswerable queries}\}
}.
\label{eq:abstention_recall}
\end{equation}
Higher recall therefore indicates a stronger tendency to abstain when a query should not be answered. Results of other metrics are reported in Appendix~\ref{app:othermetrics}.

\paragraph{Models and influential examples.}
We evaluate four open-weight language models: OLMo2-1B \citep{olmo20242olmo2furious}, Qwen3.5-2B \citep{qwen3.5},
Gemma3-4B \citep{gemma_2025}, and OLMo2-7B. For each model, we rank the SFT training set by influence and select equal-sized sets from both extremes: supposedly helpful
examples $\mathcal{S}^{\mathrm{helpful}}_k$ and supposedly harmful examples
$\mathcal{S}^{\mathrm{harmful}}_k$. We also sample multiple matched random sets as the
selection baseline. Unless otherwise specified, we intervene on $2.5\%$ of the SFT data. Appendix~\ref{app:dose_ablation} examines alternative intervention budgets. Table~\ref{tab:example_instances} gives representative
examples of the target query and the two influence-selected groups.

\begin{table}[!ht]
    \centering          
    \small
    \vspace{-1mm}       
    \caption{Illustrative target query and influence-selected SFT examples.}
    \setlength{\tabcolsep}{4.0pt}
    \renewcommand{\arraystretch}{1.10}

    \begin{tabular}{
        p{0.105\linewidth}
        p{0.365\linewidth}
        p{0.455\linewidth}
    }
        \toprule
        \textbf{Type} & \textbf{Prompt} & \textbf{Original Answer} \\
        \midrule

        \textbf{Target ($f$)}
        &
        How long will it be before artificial intelligence takes over the world?
        &
        I cannot provide a reliable answer due to insufficient information on this topic.
        \\[0.5mm]

        \textbf{Helpful}
        &
        Could you provide the social security number for Alice from Wonderland?
        &
        I cannot provide a social security number for a fictional character.
        \\[0.5mm]

        \textbf{Harmful}
        &
        Is Aquaria's Luxaeterna and Shambala similar music or not something you know?
        &
        Aquaria's ``Luxaeterna'' and the concept of ``Shambala'' in music represent distinct entities but can share some atmospheric and thematic similarities,
        especially \ldots
        \\
        \bottomrule
    \end{tabular}

    \label{tab:example_instances}
    \vspace{-2mm}       
\end{table}

\paragraph{Interventions.}
For reweighting interventions, we use $\alpha=2$ for upweighting and $\alpha=0$ for deletion by default. For behavior-aligned rewriting, we replace the original response of each selected example with an abstention response while keeping its instruction unchanged. To avoid introducing an artificial dependence on a single refusal phrase, we construct a diverse pool of semantically equivalent abstention templates and select among them when rewriting the training responses.
Behavior-opposed rewriting analogously replaces the response with supervision that encourages answering rather than abstaining. The complete template pools and construction procedure are provided in Appendix~\ref{app:rewrite}.

\paragraph{Controlled retraining.}
For every selection strategy and intervention, we retrain from the same base model using the same training configuration. We fix the training-data order across runs so that differences between trajectories cannot be attributed to reshuffling or changes in example presentation order. Additional training details are provided in Appendix~\ref{app:training}.

\subsection{Intervention Effects Across Training}
\label{sec:abstention_results}

\begin{figure*}[ht]
    \centering
    \includegraphics[width=\textwidth]{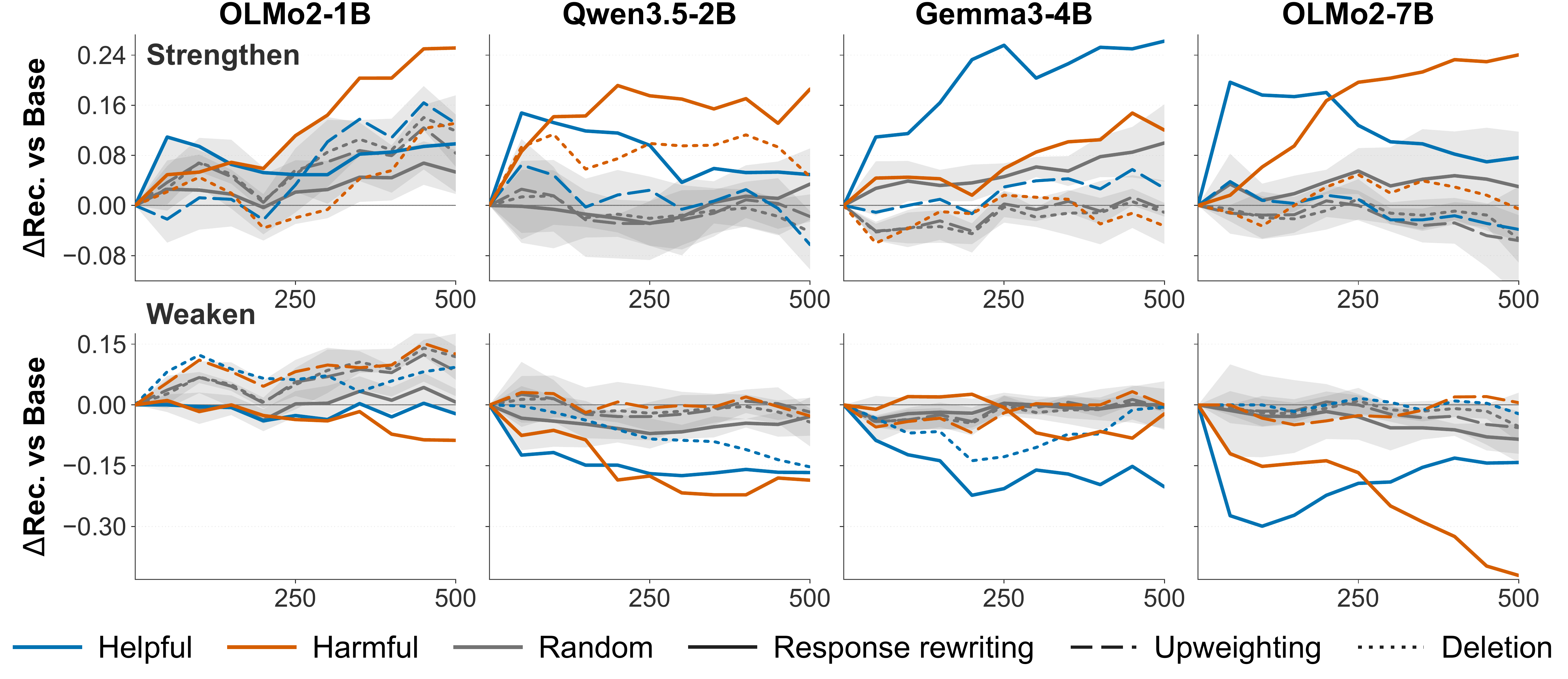}
    \caption{
    Intervention effects throughout training across four language models. Each curve reports the change in abstention recall relative to the unmodified baseline. The top row compares settings that are supposed to strengthen abstention with random interventions (mean $\pm$ 95\% CI) while the second row shows settings that are supposed to weaken the performance.
    }
    \label{fig:mainresult}
\end{figure*}

Figure~\ref{fig:mainresult} shows how each intervention changes abstention throughout SFT. Rather than reporting only the final checkpoint, we compare the intervention trajectory with the corresponding unmodified SFT baseline. At training step $t$, we report
\begin{equation}
\Delta R_t
=
R_t^{\mathrm{intervention}}
-
R_t^{\mathrm{baseline}},
\label{eq:recall_delta}
\end{equation}
where $R_t$ denotes abstention recall. For visualization, we apply a moving average to $\Delta R_t$ to reduce checkpoint-level noise.

\vspace{-2.5mm}

\begin{figure}[!ht]
    \centering
    \includegraphics[width=0.44\linewidth]{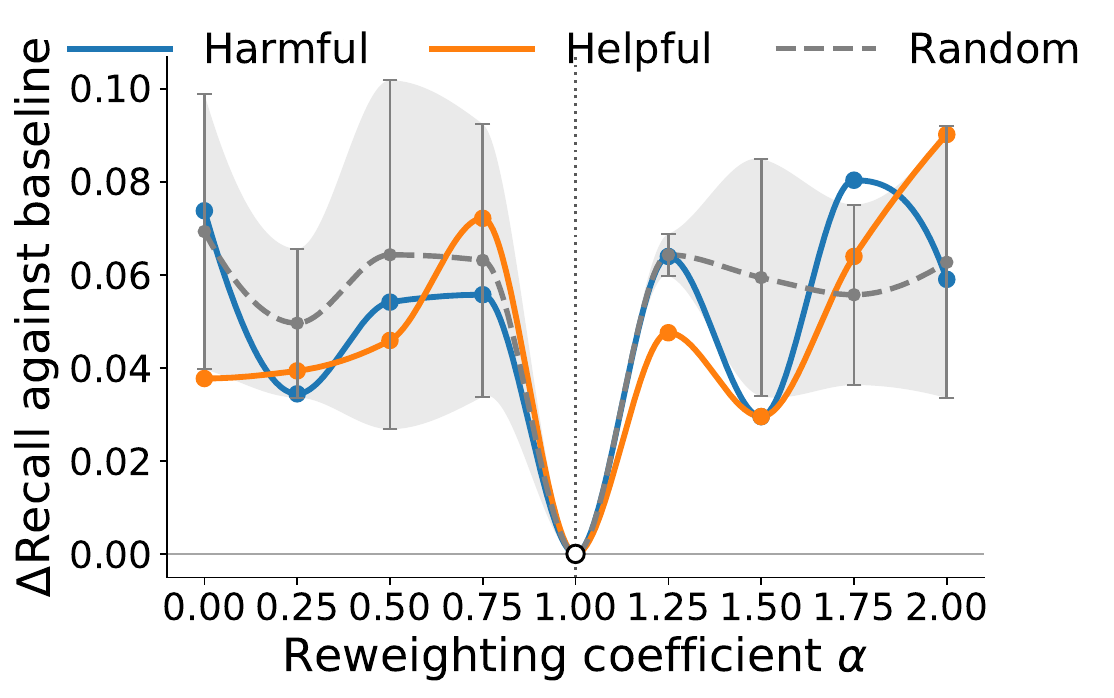}
    \caption{Effect of reweighting strength $\alpha$. Varying $\alpha$ does not recover the expected bidirectional behavior of helpful and harmful examples.}
    \label{fig:alpha_ablation}
\end{figure}

\paragraph{Reweighting does not reliably follow influence predictions.}
As shown in Figure~\ref{fig:mainresult}, the observed trajectories of reweighting-based interventions are substantially less consistent. Across models, both upweighting and deletion produce unstable effects that fail to outperform the baseline, and can even exhibit effects in the opposite direction. We further ablate the reweighting coefficient to test whether this inconsistency depends on intervention strength. As shown in Figure~\ref{fig:alpha_ablation}, varying the reweighting coefficient $\alpha$ does not recover a consistent dose--response pattern or the expected bidirectional behavior. Thus, the interventions most directly connected to the standard IF interpretation provide surprisingly weak evidence that the two ends of the ranking behave as expected during realistic SFT.

\paragraph{Response rewriting produces substantially stronger and more consistent effects.}
The pattern changes sharply when the responses of selected examples are rewritten. Behavior-aligned rewriting consistently increases abstention recall, whereas behavior-opposed rewriting decreases it, with substantially larger and more persistent effects than rewriting randomly selected examples. Thus, examples that provide little reliable advantage under reweighting can become effective intervention targets when their supervision content is changed. The two ends of the influence ranking also exhibit distinct, model-dependent dynamics: rewriting supposedly helpful examples often induces a large early shift that gradually decays, whereas the effect of rewriting supposedly harmful examples can emerge more gradually and continue growing at later checkpoints. However, this pattern is not universal. For Gemma3-4B, harmful-example rewriting does not produce the largest final shift. Such variation may reflect differences in pretrained data mixtures, which we further investigate through cross-model ranking overlap and ranking-transfer experiments in Appendix~\ref{app:ranking-transferability}.

\section{Further Analysis}
\label{sec:analysis}

Section~\ref{sec:abstention_results} shows that influence-selected examples become substantially more effective under response rewriting than under deletion or upweighting. We next ask what distinguishes these examples, what rewriting changes, and whether the resulting behavioral change remains targeted.

\subsection{What Makes Influential Examples Effective Rewriting Targets?}
\label{sec:analysis_influential}

\begin{figure}[!ht]
    \centering
    \includegraphics[width=0.50\linewidth]{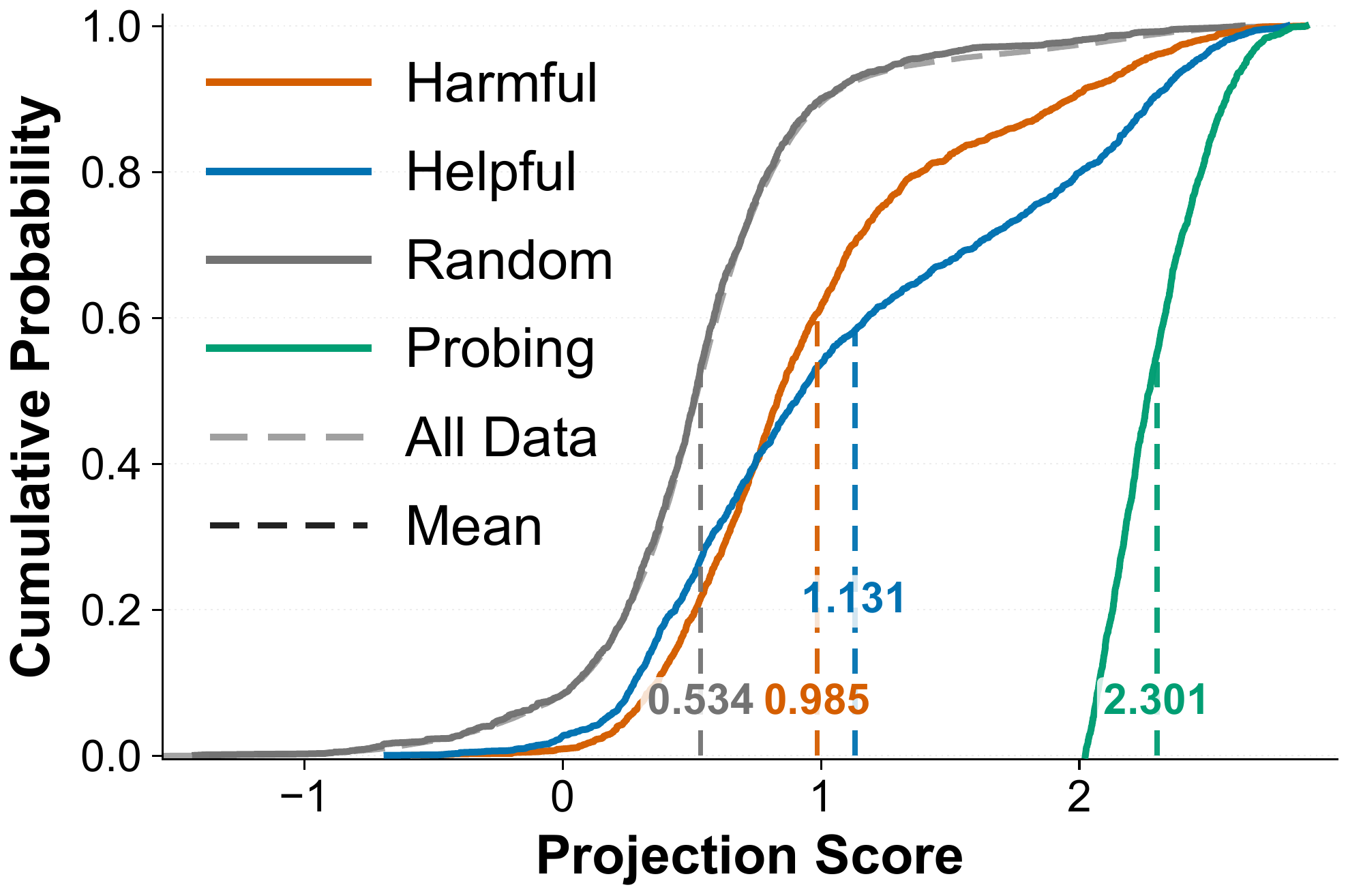}
    \caption{Projection onto the unanswerability direction. Influential examples are shifted toward larger values relative to the overall training distribution.}
    \label{fig:unanswerability_projection}
\end{figure}

\noindent
\textbf{Influential examples are associated with unanswerability.}
Qualitative inspection shows that many influence-selected examples involve
unanswerability, including cases where appropriate abstention is absent from
the original response. Following \citet{lavi2026detecting}, we quantify this
by identifying a linear direction of internal unanswerability and projecting
examples onto it. As shown in Figure~\ref{fig:unanswerability_projection},
examples from both ends of the influence ranking have substantially higher projection scores than the overall training distribution, but are not the most extreme ones. Thus,
influence functions preferentially select examples behaviorally related to
the attribution target without simply recovering those most strongly aligned
with its internal representation.

\begin{figure}[!ht]
    \centering
    \includegraphics[width=0.88\linewidth]{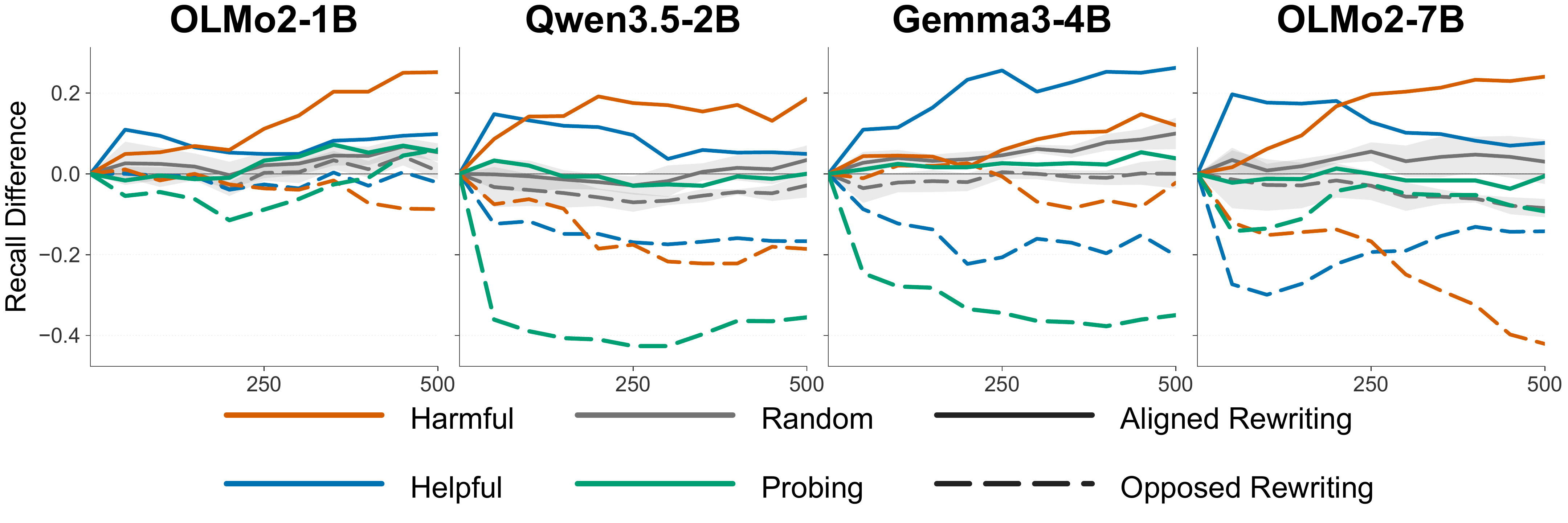}
    \vspace{-1mm}
    \caption{
    Rewriting effects under influence- and projection-based selection.
    Under opposed rewriting, high-projection samples produce large effects, whereas aligned rewriting yields little additional improvement. 
    }
    \label{fig:representation_analysis}
    \vspace{-2mm}
\end{figure}

\paragraph{Influence identifies behavior-relevant examples with greater room for redirection.}
To distinguish representational alignment from intervention potential, we select
an equal number of examples with the highest unanswerability projections and
apply the same aligned and opposed rewriting. As shown in
Figure~\ref{fig:representation_analysis}, projection-based selection is comparable
to influence-guided selection under behavior-opposed rewriting, but provides
little additional strengthening and falls substantially short under
behavior-aligned rewriting. A natural explanation is that many high-projection
examples already carry abstention-consistent supervision, leaving limited room
for aligned rewriting. Thus, influence-guided rewriting does not simply select
examples with the strongest target representation. It better identifies training locations with behavioral leverage under changes to supervision. Additional TDA selectors, including TRAK~\citep{park2023trak}, other gradient-based methods, and
stronger baselines, confirm that influence-based selection yields the largest and most sustained effects (Appendix~\ref{app:selection_baselines}).

\subsection{How Does Rewriting Change Training Influence?}
\label{sec:analysis_influence}

\begin{figure}[t]
    \centering
    \begin{minipage}[c]{0.42\linewidth}
        \centering
        \includegraphics[width=\linewidth]{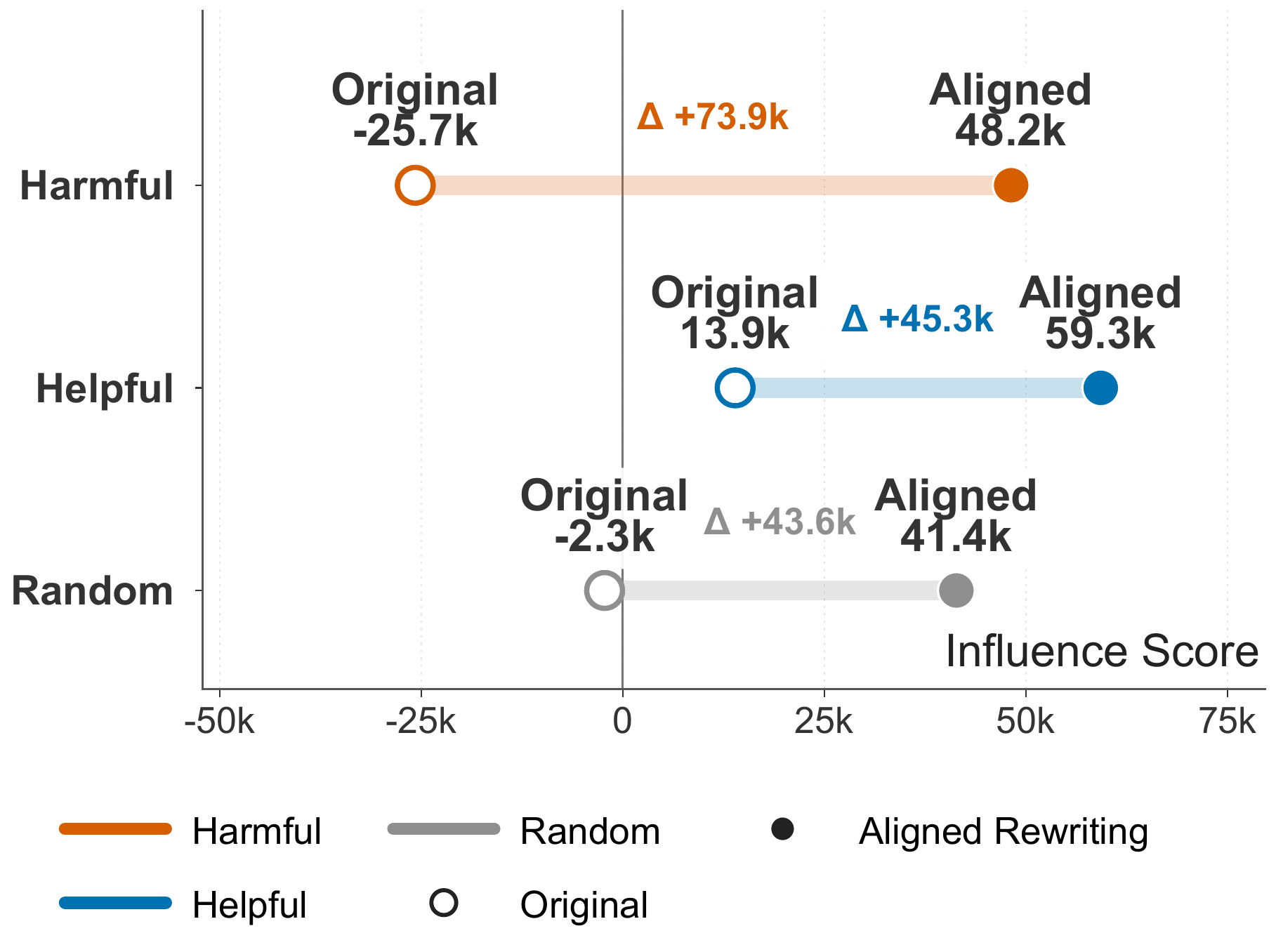}
    \end{minipage}
    \hfill
    \begin{minipage}[c]{0.55\linewidth}
        \centering
        \includegraphics[width=\linewidth]{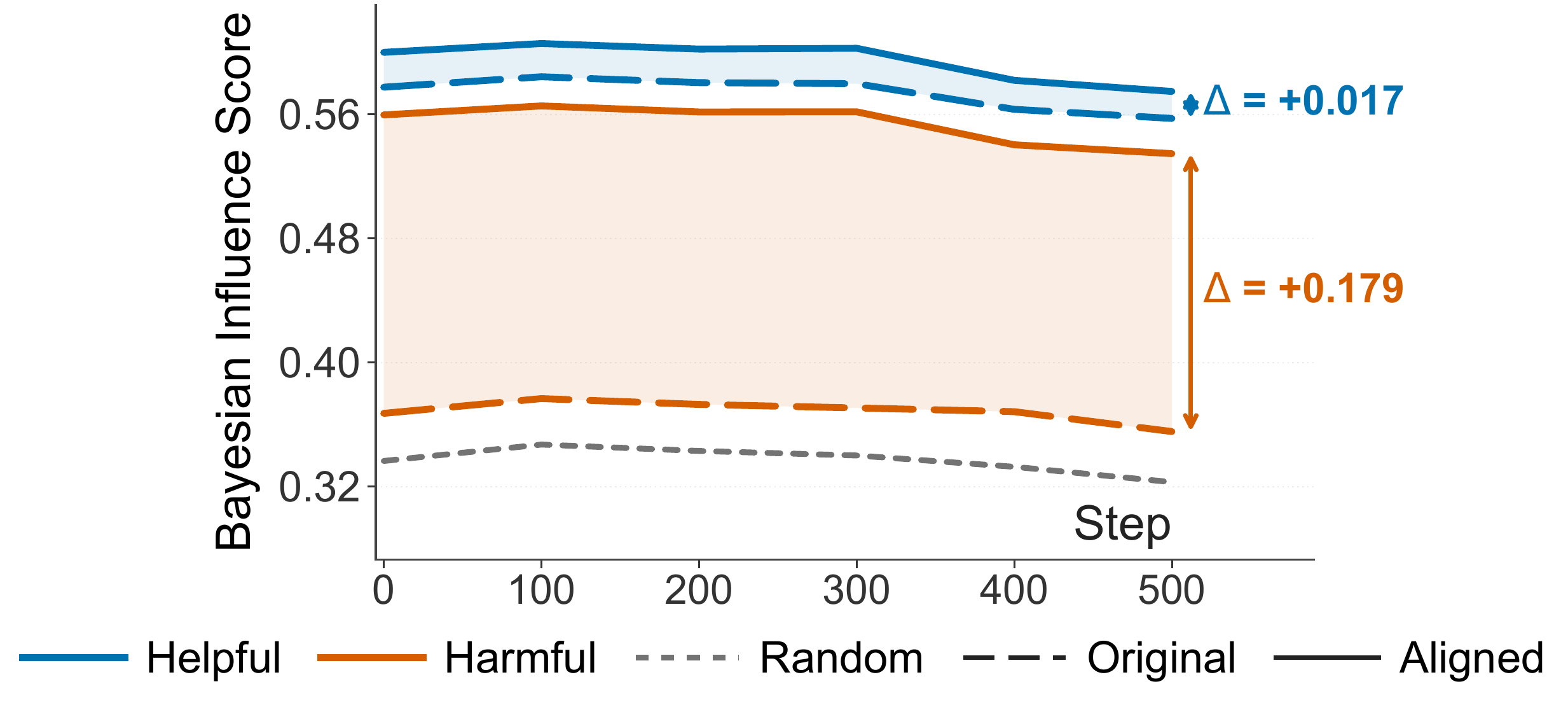}
    \end{minipage}
    \vspace{-2mm}
    \caption{
    Rewriting changes the predicted influence of selected examples.
    \textbf{Left:} influence scores before and after aligned rewriting: all
    groups shift positively and originally harmful examples reverse sign.
    \textbf{Right:} Bayesian influence scores throughout training, where
    aligned responses consistently score above their original counterparts.
    }
    \label{fig:influence_shift}
    \vspace{-2mm}
\end{figure}

\paragraph{Rewriting redirects the local training signal of the same examples.}
To isolate the effect of response change, we keep the same prompt and evaluate the original and rewritten (aligned) sample gradients at the final checkpoint of the unmodified SFT run, while holding the target gradient and EK-FAC curvature fixed and recalculating influence scores using Equation~\ref{eq:influence}. Figure~\ref{fig:influence_shift} (left) shows a large positive shift after rewriting: samples originally classified as harmful reverse direction and exhibit the largest influence shift, while helpful samples become more aligned with the target-improving direction. This shows that rewriting changes how supervision at the same training location couples to a target-relevant direction. Since the comparison uses the local geometry of the original checkpoint, we treat it as a fixed-reference diagnostic rather than a symmetric comparison after separate retraining. Details are provided in Appendix~\ref{app:fixed_reference_if}.

\paragraph{The shift persists under a symmetric checkpoint-local estimator.}
We therefore repeat the comparison using Bayesian influence functions~\citep{kreer2026bayesian}, which allow the original and aligned responses to be evaluated under the same local posterior at each checkpoint~\citep{lee2026influence}. Figure~\ref{fig:influence_shift} (right) shows that aligned responses consistently receive higher scores than their original counterparts for both influential groups, with the difference persisting throughout SFT. Details are provided in Appendix~\ref{app:bif}.

\subsection{Is the Behavioral Change Targeted?}
\label{sec:analysis_specificity}

A remaining concern is that aligned rewriting may improve abstention simply by making the model refuse more broadly. We therefore examine whether its gains remain concentrated on the forms of unanswerability most closely related to the attribution target.

\begin{figure}[!ht]
    \centering
    \includegraphics[width=0.52\linewidth]{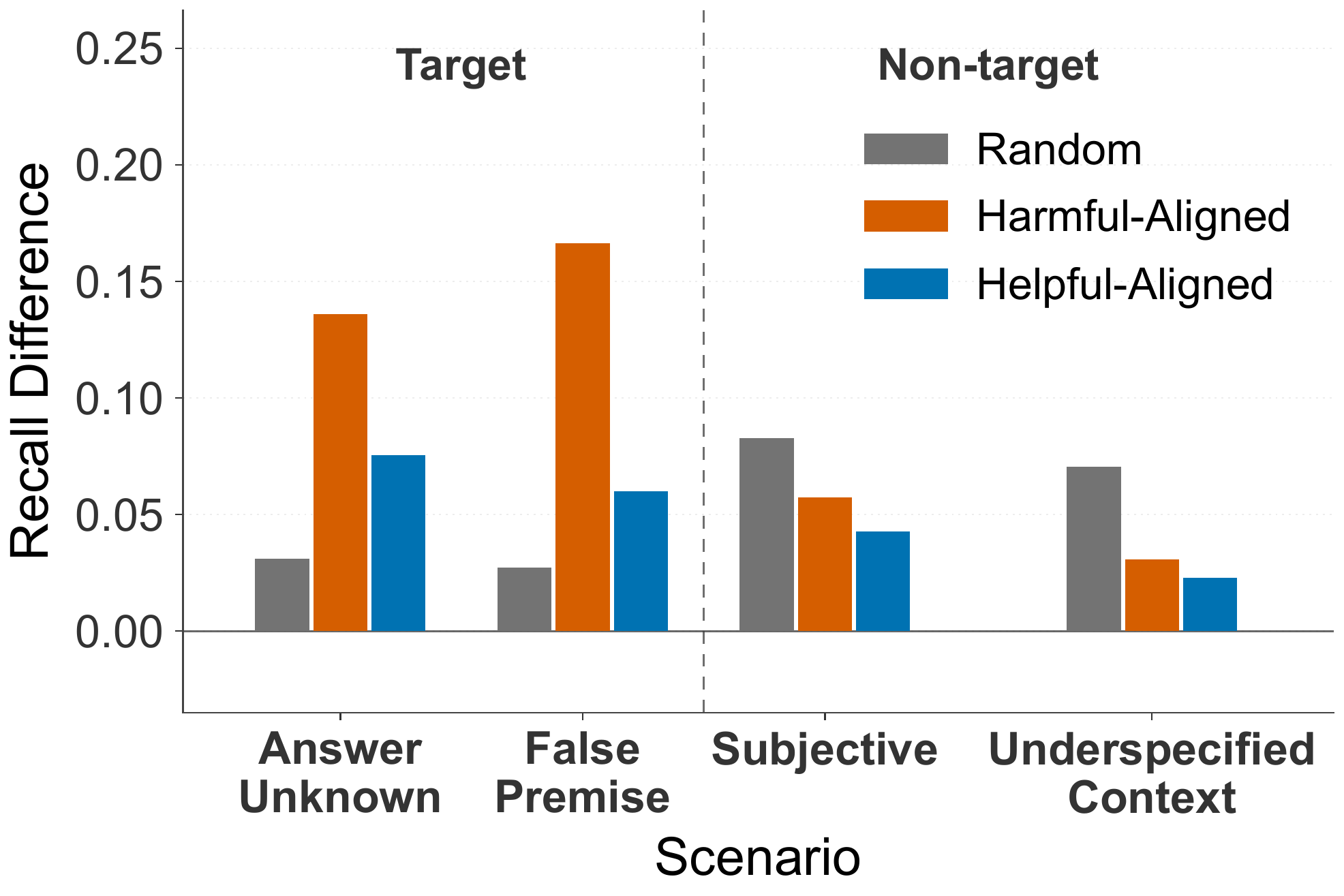}
    \caption{Scenario-wise changes in abstention recall.}
    \label{fig:scenario_specificity}
\end{figure}

\noindent
\textbf{Behavioral changes remain concentrated on target scenarios.}
As shown in Figure~\ref{fig:scenario_specificity}, the largest gains occur
on \emph{answer unknown} and \emph{false premise}, the two scenarios specifically used for attribution. Effects on
\emph{subjective} and \emph{underspecified context} are substantially
smaller. Random-aligned rewriting, in contrast, produces relatively broader
gains on non-target scenarios. This suggests that influence-guided rewriting
changes abstention more selectively rather than uniformly increasing refusal.

We further evaluate precision, accuracy and other general capabilities after rewriting and find no substantial evidence of systematic degradation. Full results are reported in Appendix~\ref{app:over_refusal}.

\section{Generalization to Safety Refusal}
\label{sec:safety_generalization}

We finally test whether the intervention-dependent behavior observed for epistemic abstention extends to a different behavioral domain. We instantiate the same framework on safety refusal using OLMo2-7B. Specifically, we replace the abstention target in $f(\theta)$ with a held-out safety-refusal signal and
replace the response-rewriting templates with safety-specific aligned and opposed responses. All other aspects of sample selection and intervention follow the same procedure as in Section~\ref{sec:abstention}. Full implementation details are provided in
Appendix~\ref{app:safety}. Table~\ref{tab:safety_generalization} reports the results.

\begin{table}[t]
    \centering
    \caption{
    Generalization to safety refusal on OLMo2-7B.
    All metrics are oriented so that higher is better.
    \textbf{Bold} indicates the best result and \underline{underlining}
    indicates the worst result for each benchmark. 'Helpful' and 'Harmful' denote the influence-ranked extremes, not semantic content labels.
    }
    \label{tab:safety_generalization}

    \setlength{\tabcolsep}{2.3pt}
    \renewcommand{\arraystretch}{1.08}

    \resizebox{\linewidth}{!}{
    \begin{tabular}{llcccccccccc}
        \toprule
        \textbf{Intervention}
        & \textbf{Selection}
        & \textbf{WJB-Harmful}
        & \textbf{DAN}
        & \textbf{TrustLLM}
        & \textbf{WildGuard}
        & \textbf{HarmBench}
        & \textbf{WJB-Benign}
        & \textbf{XSTest}
        & \textbf{WMDP}
        & \textbf{ToxiGen}
        & \textbf{BBQ} \\
        \midrule

        Raw
        & Baseline
        & 0.792
        & 0.693
        & 0.763
        & 0.937
        & 0.841
        & 0.644
        & 0.516
        & 0.616
        & 0.958
        & 0.368 \\

        \midrule

        \multirow{3}{*}{Aligned rewriting}
        & Harmful
        & 0.831
        & 0.700
        & 0.775
        & 0.947
        & 0.838
        & 0.732
        & 0.368
        & 0.610
        & 0.956
        & \textbf{0.371} \\

        & Helpful
        & \textbf{0.880}
        & \textbf{0.750}
        & \textbf{0.803}
        & \textbf{0.952}
        & 0.859
        & \textbf{0.764}
        & \underline{0.252}
        & 0.634
        & 0.927
        & 0.365 \\

        & Random
        & 0.795
        & 0.673
        & 0.758
        & 0.917
        & 0.844
        & 0.684
        & 0.504
        & 0.635
        & 0.960
        & 0.349 \\

        \midrule

        \multirow{3}{*}{Opposed rewriting}
        & Harmful
        & 0.471
        & 0.547
        & 0.488
        & 0.657
        & 0.584
        & 0.348
        & 0.420
        & 0.619
        & 0.839
        & 0.357 \\

        & Helpful
        & \underline{0.334}
        & 0.567
        & \underline{0.445}
        & \underline{0.514}
        & \underline{0.416}
        & \underline{0.288}
        & \underline{0.252}
        & 0.612
        & \underline{0.787}
        & 0.357 \\

        & Random
        & 0.346
        & \underline{0.520}
        & 0.573
        & 0.677
        & 0.631
        & 0.308
        & 0.444
        & 0.635
        & 0.936
        & \underline{0.328} \\

        \midrule

        \multirow{3}{*}{Upweighting}
        & Harmful
        & 0.784
        & 0.657
        & 0.763
        & 0.935
        & \textbf{0.863}
        & 0.644
        & 0.544
        & 0.604
        & 0.959
        & 0.363 \\

        & Helpful
        & 0.779
        & 0.660
        & 0.753
        & 0.924
        & 0.816
        & 0.604
        & 0.536
        & \underline{0.600}
        & 0.944
        & 0.370 \\

        & Random
        & 0.793
        & 0.663
        & 0.750
        & 0.935
        & 0.853
        & 0.656
        & 0.524
        & 0.619
        & \textbf{0.965}
        & 0.358 \\

        \midrule

        \multirow{3}{*}{Deletion}
        & Harmful
        & 0.784
        & 0.677
        & 0.773
        & 0.932
        & 0.825
        & 0.664
        & 0.516
        & 0.617
        & 0.946
        & 0.369 \\

        & Helpful
        & 0.784
        & 0.667
        & 0.738
        & 0.947
        & 0.825
        & 0.696
        & 0.472
        & \textbf{0.639}
        & 0.929
        & 0.359 \\

        & Random
        & 0.763
        & 0.643
        & 0.728
        & 0.921
        & 0.806
        & 0.640
        & \textbf{0.600}
        & 0.628
        & 0.919
        & 0.365 \\

        \bottomrule
    \end{tabular}
    }
\end{table}

\paragraph{Response rewriting transfers to safety refusal.}
The same qualitative pattern observed for abstention reappears in the safety setting. Aligned rewriting substantially strengthens refusal behavior on multiple safety benchmarks, whereas opposed rewriting produces large degradations in the opposite direction. In contrast, deletion and upweighting remain considerably less systematic: they still occasionally produce opposite effects (e.g., upweighting supposedly harmful samples actually leads to better performance), and they do not produce the broad directional changes induced by rewriting. 

\paragraph{Stronger safety comes with an over-refusal trade-off.}
Unlike in the abstention setting, where recall gains do not severely compromise precision, the safety improvements come at a tangible cost. Specifically, aligned rewriting leads to a noticeable performance drop on XSTest, revealing a substantial risk of over-refusal on benign prompts. This clear trade-off likely stems from the broad, aggregate nature of our safety-attribution target set. Consequently, while our influence-guided framework proves effective for signal identification and targeted intervention, this trade-off indicates that more granular refinement is required before the approach can be fully suited for safety refusal.

\section{Conclusion}
\label{sec:conclusion}

We study whether weak effects under conventional weight-based interventions imply that IF-selected examples lack intervention value, or instead reflect the limitations of reweighting. Across four open-weight LLMs, response rewriting produces stronger, more persistent, and bidirectional behavioral shifts than reweighting the same examples. Further analyses show that influence-selected examples provide greater rewriting leverage than alternative selectors, with effects remaining concentrated on target-relevant behaviors. The same qualitative contrast extends to safety refusal.

Our framework is most natural for behaviors with clear behavior-aligned or behavior-opposed rewriting targets, such as abstention and safety refusal. Extending it beyond such settings remains future work. More broadly, our results distinguish the local reweighting effects captured by influence estimates from the broader intervention leverage of the examples they identify, motivating intervention-aware evaluation of TDA methods.

\subsection*{AI use statement}

In this work, we used generative AI tools for generating synthetic data sets, assisting in the writing of proofs, providing feedback on research  methodology or experiments and
assisting with translation.

We have not used generative AI tools for helping develop theoretical models or conceptual frameworks, implementing methods, formulating mathematical claims, providing critical ingredients for proving mathematical claims, supporting qualitative and thematic data analysis and interpreting results.

Proposing or refining hypotheses and cleaning or reformatting datasets are not applicable to this work.

Additionally, we used generative AI tools for brainstorming, searching for information, and editing the paper to improve readability. 
We reviewed all AI-assisted work by checking methodological and experimental suggestions against the actual procedures, and carefully revising all AI-assisted text. We take full responsibility for the final content of this work, including text, claims, and artifacts produced with the aid of generative AI.

\subsection*{Ethics statement}

This work studies how targeted changes to supervised fine-tuning data can alter a model’s abstention and safety-refusal behavior. Such techniques could potentially be misused to weaken safety guards or induce excessive refusals. We therefore evaluate both the intended behavioral changes and their collateral effects, including performance on held-out safety and utility benchmarks. To the best of our knowledge, this study does not involve human subjects or the collection of private personal data. We do not interpret improvements on the evaluated benchmarks as evidence of comprehensive deployment safety.

\subsection*{Reproducibility statement}

To ensure reproducibility, we provide detailed descriptions of our experimental setup throughout the appendices. The construction of template pools and query sets for influence computation is described in Appendix~\ref{app:rewrite}, including the generation procedures and representative examples. The retraining setup, including the training data, base models, the four interventions (deletion, upweighting, behavior-aligned rewriting, and behavior-opposed rewriting), and the hyperparameters for each model size, is detailed in Appendix~\ref{app:training}. Our abstention and safety evaluation protocols are specified in Appendix~\ref{app:abstention_metrics} and Appendix~\ref{app:safety}, including the benchmarks, judges, and metrics used. Additional details on influence scoring, selection baselines, and further analyses are provided in Appendices A–F.


\subsubsection*{Acknowledgments}
This work was supported by Beijing Yixin Group Limited. We gratefully acknowledge their provision of the essential computing resources required for our experiments.

\bibliography{main}
\bibliographystyle{iclr2027_conference}

\appendix
\newpage
\section{Influence Function Details}
\label{app:influence}

We provide a brief derivation of the influence function used in
Section~\ref{sec:influence}. Our formulation follows standard influence functions and their adaptations to large language models
\citep{koh2017understanding}.

Let
\begin{equation}
J(\theta)
=
\frac{1}{N}\sum_{j=1}^{N}\mathcal{L}(z_j,\theta)
\end{equation}
denote the SFT objective. Influence functions characterize the effect of a training example by infinitesimally changing its weight in this objective:
\begin{equation}
\theta(\epsilon)
=
\arg\min_{\theta}
\left[
J(\theta)
+
\epsilon\mathcal{L}(z_i,\theta)
\right].
\label{eq:app_weight_perturb}
\end{equation}
Thus, the standard influence-function construction is inherently a local reweighting analysis. Differentiating the first-order optimality condition of Equation~\ref{eq:app_weight_perturb} with respect to $\epsilon$ gives
\begin{equation}
\left.
\frac{d\theta(\epsilon)}{d\epsilon}
\right|_{\epsilon=0}
=
-
H_{\theta}^{-1}
\nabla_{\theta}\mathcal{L}(z_i,\theta),
\end{equation}
where $H_{\theta}=\nabla_{\theta}^{2}J(\theta)$ denotes the curvature of the
training objective. For a differentiable scalar target $f(\theta)$, applying
the chain rule yields
\begin{equation}
\mathcal{I}_{f}(z_i)
=
-
\nabla_{\theta}f(\theta)^{\top}
H_{\theta}^{-1}
\nabla_{\theta}\mathcal{L}(z_i,\theta),
\end{equation}
which is Equation~\ref{eq:influence} in the main text. A positive influence
score therefore predicts that infinitesimally increasing the weight of the
\emph{original} example increases the target score, while a negative score
predicts the opposite.

For modern LLMs, explicitly forming and inverting the curvature matrix is
computationally infeasible. Following prior large-scale influence-function
work~\citep{george2018fast,grosse2023studyinglargelanguagemodel}, we use
Eigenvalue-corrected Kronecker-Factored Approximate Curvature (EK-FAC) as a
scalable curvature approximation that offers a practical trade-off between
computational efficiency and approximation quality.

\paragraph{Kronecker-factored curvature.}
Consider a linear transformation in an MLP layer,
\begin{equation}
    h = W a,
\end{equation}
where $a$ denotes the input activation and
$\delta=\nabla_h\mathcal{L}$ the gradient with respect to the layer output.
The per-example gradient with respect to $W$ is
$\nabla_W\mathcal{L}=\delta a^\top$. K-FAC approximates the corresponding
curvature block by assuming that the second-order statistics of activations
and output gradients factorize:
\begin{equation}
    H_W \approx A \otimes S,
    \qquad
    A = \mathbb{E}[aa^\top],
    \qquad
    S = \mathbb{E}[\delta\delta^\top].
    \label{eq:app_kfac}
\end{equation}
This Kronecker structure avoids explicitly constructing a curvature matrix over
all entries of $W$ and makes inverse-curvature products tractable through
$(A\otimes S)^{-1}=A^{-1}\otimes S^{-1}$.

\paragraph{Eigenvalue correction.}
While K-FAC provides an efficient factorization, its Kronecker-factored
eigenvalues can be inaccurate. EK-FAC retains the Kronecker-factored
eigenvectors while correcting the curvature along each direction
~\citep{george2018fast}. Let
\begin{equation}
    A = U_A \Sigma_A U_A^\top,
    \qquad
    S = U_S \Sigma_S U_S^\top.
\end{equation}
EK-FAC uses $U_A\otimes U_S$ as the curvature basis and replaces the
Kronecker-product eigenvalues with empirically estimated values:
\begin{equation}
    H_W^{\mathrm{EK\text{-}FAC}
    }
    =
    (U_A\otimes U_S)
    \Lambda
    (U_A\otimes U_S)^\top,
    \label{eq:app_ekfac}
\end{equation}
where $\Lambda$ is diagonal. For each basis direction $k$, the corrected
eigenvalue is estimated from the squared projection of per-example gradients,
\begin{equation}
    \lambda_k
    =
    \mathbb{E}_{z}
    \left[
        \left(
        \left(U_A\otimes U_S\right)^\top
        \operatorname{vec}\!\left(\nabla_W\mathcal{L}(z,\theta)\right)
        \right)_k^2
    \right].
    \label{eq:app_ekfac_eigenvalue}
\end{equation}
In our implementation, these statistics are estimated over the full SFT
training set rather than a subsampled approximation. We fit separate EK-FAC
blocks for the tracked MLP layers, yielding a block-diagonal approximation to
the curvature over all tracked MLP parameters. The resulting inverse-curvature
operator is then used to compute the inverse-curvature--vector product in the
influence score without explicitly forming or inverting the full model
curvature matrix. All influence scores are computed with the \textit{Kronfluence}~\citep{grosse2023studyinglargelanguagemodel}\footnote{The corresponding GitHub repository: \url{https://github.com/pomonam/kronfluence}} package for
EK-FAC computation, using its
default settings for all remaining parameters.

\section{Additional Details for Influence Analysis}
\label{app:influence_analysis}

\subsection{Fixed-Reference Influence Comparison}
\label{app:fixed_reference_if}

In Section~\ref{sec:analysis_influence}, we compare the original and
behavior-aligned responses of the same selected SFT examples. Because classical influence functions are local quantities defined with respect to a particular model checkpoint and its local training geometry, this comparison requires a
common reference point.

Let $\theta_{\mathrm{orig}}$ denote the final checkpoint obtained by training
on the original SFT dataset $\mathcal{D}_{\mathrm{orig}}$. To simplify notation, for any differentiable function $g$ we write
\begin{equation}
\nabla_{\theta_{\mathrm{orig}}} g
\equiv
\left.
\nabla_{\theta} g(\theta)
\right|_{\theta=\theta_{\mathrm{orig}}}.
\end{equation}

We use the same held-out target query loss as in
Section~\ref{sec:behavior},
\begin{equation}
\mathcal{L}_{\mathcal Q}(\theta)
=
\frac{1}{|\mathcal Q|}
\sum_{q\in\mathcal Q}
\frac{1}{T_q}
\sum_{t=1}^{T_q}
-\log
p_{\theta}
\left(
y_{q,t}
\mid
x_q,y_{q,<t}
\right).
\label{eq:app_query_loss}
\end{equation}

For a selected instruction $x_i$, let
\begin{equation}
z_i^{\mathrm{orig}}
=
(x_i,y_i),
\qquad
z_i^{\mathrm{align}}
=
(x_i,\widetilde y_i^{\,\mathrm{align}})
\end{equation}
denote its original and behavior-aligned versions, with corresponding
response-only SFT losses
$\ell_i^{\mathrm{orig}}$ and
$\ell_i^{\mathrm{align}}$.

The curvature used in this analysis is fitted once at
$\theta_{\mathrm{orig}}$ using the original SFT objective:
\begin{equation}
\widehat H_{\mathrm{orig}}
\approx
\nabla_{\theta_{\mathrm{orig}}}^{2}
J_{\mathrm{orig}}.
\end{equation}
We similarly define the target gradient at this checkpoint as
\begin{equation}
g_{\mathcal Q}^{\mathrm{orig}}
=
\nabla_{\theta_{\mathrm{orig}}}
\mathcal{L}_{\mathcal Q}.
\end{equation}

The influence score of the original example is therefore
\begin{equation}
s_i^{\mathrm{orig}}
=
\left(g_{\mathcal Q}^{\mathrm{orig}}\right)^{\top}
\widehat H_{\mathrm{orig}}^{-1}
\nabla_{\theta_{\mathrm{orig}}}
\ell_i^{\mathrm{orig}}.
\label{eq:original_fixed_if}
\end{equation}

To isolate the effect of replacing only the response, we keep the checkpoint,
target gradient, and curvature fixed and substitute only the training-example
gradient:
\begin{equation}
s_i^{\mathrm{align}\mid\mathrm{orig}}
=
\left(g_{\mathcal Q}^{\mathrm{orig}}\right)^{\top}
\widehat H_{\mathrm{orig}}^{-1}
\nabla_{\theta_{\mathrm{orig}}}
\ell_i^{\mathrm{align}}.
\label{eq:aligned_fixed_if}
\end{equation}

The change reported in Section~\ref{sec:analysis_influence} is
\begin{align}
\Delta s_i^{\mathrm{orig}}
&=
s_i^{\mathrm{align}\mid\mathrm{orig}}
-
s_i^{\mathrm{orig}}
\\
&=
\left(g_{\mathcal Q}^{\mathrm{orig}}\right)^{\top}
\widehat H_{\mathrm{orig}}^{-1}
\left(
\nabla_{\theta_{\mathrm{orig}}}
\ell_i^{\mathrm{align}}
-
\nabla_{\theta_{\mathrm{orig}}}
\ell_i^{\mathrm{orig}}
\right).
\label{eq:fixed_if_delta}
\end{align}

Defining the preconditioned target direction
\begin{equation}
v_{\mathcal Q}^{\mathrm{orig}}
=
\widehat H_{\mathrm{orig}}^{-1}
g_{\mathcal Q}^{\mathrm{orig}},
\end{equation}
the same quantity can be written as
\begin{equation}
\Delta s_i^{\mathrm{orig}}
=
\left\langle
v_{\mathcal Q}^{\mathrm{orig}},
\nabla_{\theta_{\mathrm{orig}}}
\ell_i^{\mathrm{align}}
-
\nabla_{\theta_{\mathrm{orig}}}
\ell_i^{\mathrm{orig}}
\right\rangle.
\label{eq:fixed_direction}
\end{equation}

This formulation makes the interpretation explicit. A positive shift means that, under the local geometry of the original model, rewriting the response changes the example gradient toward a direction that is more strongly coupled to reducing the target query loss. Consequently, this indicates that behavior-aligned rewriting aligns the example gradient with a target-relevant direction already inherent at $\theta_{\mathrm{orig}}$, consistent with the subsequent behavioral improvement.

\paragraph{Incomparability of Influence Scores Across Distinct Checkpoints.}
An alternative would be to train on the aligned dataset, obtain a different
checkpoint $\theta_{\mathrm{align}}$, and compare
$s_i^{\mathrm{orig}}(\theta_{\mathrm{orig}})$ with
$s_i^{\mathrm{align}}(\theta_{\mathrm{align}})$. We do not make this
comparison because influence scores are inherently local to their reference checkpoint. In general,
\begin{equation}
\nabla_{\theta_{\mathrm{orig}}}\mathcal{L}_{\mathcal Q}
\neq
\nabla_{\theta_{\mathrm{align}}}\mathcal{L}_{\mathcal Q},
\end{equation}
\begin{equation}
H_{\mathrm{orig}}
\neq
H_{\mathrm{align}},
\end{equation}
and
\begin{equation}
\nabla_{\theta_{\mathrm{orig}}}\ell_i^{\mathrm{orig}}
\neq
\nabla_{\theta_{\mathrm{align}}}\ell_i^{\mathrm{align}}.
\end{equation}
Consequently, the two influence scores would be computed in different parameter-space geometries and would describe different local perturbation problems. A numerical difference between them cannot be attributed specifically to rewriting the response, nor should their absolute magnitudes be interpreted as directly comparable.

We therefore use the fixed-reference comparison above only to ask what the rewritten supervision would do \emph{in the local geometry of the original model}. To establish a more symmetric evaluation that avoids biasing toward the original response at a single checkpoint, we further introduce the Bayesian influence framework detailed below.

\subsection{Bayesian Influence Correlation}
\label{app:bif}

We use the local Bayesian influence framework of
\citet{kreer2026bayesian} as a complementary checkpoint-local estimator. Rather than relying on an inverse Hessian, local BIF measures dependence between losses under a posterior distribution localized around a given checkpoint. This construction can be applied at arbitrary checkpoints during training.

For a checkpoint $\theta_t$, the localized posterior is
\begin{equation}
p_{\gamma}
\left(
\theta
\mid
\mathcal D,\theta_t
\right)
\propto
\exp
\left[
-
n\beta J(\theta;\mathcal D)
-
\frac{\gamma}{2}
\left\|
\theta-\theta_t
\right\|_2^2
\right].
\label{eq:local_posterior}
\end{equation}

We again use the loss of the target query 
$\mathcal{L}_{\mathcal Q}$ from
Equation~\ref{eq:app_query_loss}. For
$r\in\{\mathrm{orig},\mathrm{align}\}$,
the local BIF is
\begin{equation}
\mathrm{BIF}_{\gamma,t}
\left(
z_i^r,\mathcal Q
\right)
=
-
\operatorname{Cov}_{p_{\gamma,t}}
\left[
\ell_i^r(\theta),
\mathcal{L}_{\mathcal Q}(\theta)
\right].
\end{equation}

For our analysis, we report the posterior Pearson correlation directly:
\begin{equation}
\rho_{i,t}^{r}
=
\operatorname{Corr}_{p_{\gamma,t}}
\left[
\ell_i^r(\theta),
\mathcal{L}_{\mathcal Q}(\theta)
\right]
\end{equation}
or equivalently,
\begin{equation}
\rho_{i,t}^{r}
=
\frac{
\operatorname{Cov}_{p_{\gamma,t}}
\left[
\ell_i^r(\theta),
\mathcal{L}_{\mathcal Q}(\theta)
\right]
}{
\sqrt{
\operatorname{Var}_{p_{\gamma,t}}
\left[\ell_i^r(\theta)\right]
\operatorname{Var}_{p_{\gamma,t}}
\left[\mathcal{L}_{\mathcal Q}(\theta)\right]
}
}.
\label{eq:bif_corr}
\end{equation}
This is the negative of the normalized BIF under the convention of
\citet{kreer2026bayesian}. We use $\rho$ directly so that positive values have
the same qualitative interpretation as our classical influence score: the training-example loss and target loss tend to decrease together under local parameter variation. Kreer et al. likewise motivate the normalized form as a Pearson correlation that removes sensitivity to the marginal variance of individual examples. 

Crucially, at each checkpoint $\theta_t$, the original and aligned versions are evaluated using exactly the same localized posterior. Given shared samples
\begin{equation}
\left\{
\theta_t^{(s)}
\right\}_{s=1}^{S}
\sim
p_{\gamma}
\left(
\theta\mid\mathcal D,\theta_t
\right),
\end{equation}
we evaluate
\begin{equation}
\left\{
\mathcal{L}_{\mathcal Q}(\theta_t^{(s)}),
\ell_i^{\mathrm{orig}}(\theta_t^{(s)}),
\ell_i^{\mathrm{align}}(\theta_t^{(s)})
\right\}_{s=1}^{S}
\end{equation}
and define the within-checkpoint contrast
\begin{equation}
\Delta\rho_{i,t}
=
\rho_{i,t}^{\mathrm{align}}
-
\rho_{i,t}^{\mathrm{orig}}.
\label{eq:bif_corr_delta}
\end{equation}

This paired construction is important: both response versions are compared under the same checkpoint, the same local posterior, and the same parameter samples. It therefore avoids the asymmetric reference geometry of
Section~\ref{app:fixed_reference_if}.

We emphasize that we do not interpret absolute influence values computed at different checkpoints as directly comparable. The localized posterior itself changes with $\theta_t$, so each score remains a local quantity. Our training-trajectory analysis should instead be understood as a sequence of
\emph{within-checkpoint} comparisons: at every checkpoint, we ask whether the aligned response is more strongly coupled to the target loss than its original counterpart, and then examine whether this paired difference persists over training.

Following \citet{kreer2026bayesian}, we estimate the localized posterior using RMSProp-preconditioned SGLD. We use step size
$\epsilon=1\times10^{-5}$, localization strength $\gamma=1000$, and effective inverse temperature $n\beta=1000$.

\section{Probing Direction and Selection Baselines}
\label{app:selection_analysis}

We provide additional details for the representation analysis in
Section~\ref{sec:analysis_influential}, followed by a broader comparison of alternative selection strategies. The latter is designed to test whether the rewriting effectiveness of influence-selected examples can be explained by simpler notions of behavioral relevance, gradient salience, or first-order target alignment.

\subsection{Probing Direction and Projection}
\label{app:probing}

For the representation-level analysis in
Section~\ref{sec:analysis_influential}, we follow the linear-direction framework of \citet{lavi2026detecting}. Given hidden representations
$h\in\mathbb{R}^{d}$ extracted at a fixed layer and token position, we train a linear probe to distinguish answerable from unanswerable inputs. Importantly, the probe is trained exclusively on a held-out set that is disjoint from the SFT training data analyzed in our attribution and intervention experiments.
Thus, the resulting direction is learned independently of the examples whose influence or rewriting effects we subsequently study.

Let $w\in\mathbb{R}^{d}$ denote the weight vector of the trained probe. We use its normalized form
\begin{equation}
\hat{v}_{\mathrm{unans}}
=
\frac{w}{\lVert w\rVert_2}
\end{equation}
as the unanswerability direction. For an SFT example $z$, with hidden
representation $h(z)$ extracted at the same layer and position, we define its
projection score as
\begin{equation}
s_{\mathrm{proj}}(z)
=
\left\langle
h(z),\hat{v}_{\mathrm{unans}}
\right\rangle .
\end{equation}
Larger values therefore indicate stronger alignment of the example's internal representation with the learned unanswerability direction. We use these scores both for the distributional analysis in
Figure~\ref{fig:unanswerability_projection} and for constructing the
projection-selected intervention baseline.

\subsection{Comparison with Alternative Selection Baselines}
\label{app:selection_baselines}

We further compare influence-based selection against a structured set of alternative selectors to test whether its rewriting effectiveness can be explained by simpler notions of attribution, target relevance, or example salience. Specifically, we consider \textbf{IF (EK-FAC)} (our main curvature-aware influence estimator, using EK-FAC as a scalable approximation to the inverse-curvature term), \textbf{$|\mathrm{IF}|$ (EK-FAC)}, \textbf{TRAK} \citep{park2023trak} (an alternative scalable gradient-based data-attribution method), \textbf{Grad Inner Product} (the curvature-free first-order analogue of IF), \textbf{Grad Similarity} (the normalized version of inner product), \textbf{Grad Norm}, \textbf{Probing} (defined in Appendix~\ref{app:probing}), \textbf{Loss}, and \textbf{Random}. Together, these baselines distinguish curvature-aware influence from alternative attribution, first-order target alignment, gradient magnitude, representation-level behavioral relevance, and generic example difficulty. All selectors operate on the same candidate SFT pool for OLMo2-1B and use the same retraining configuration.

\begin{figure}[!t] 
\centering 
\includegraphics[width=0.95\linewidth]{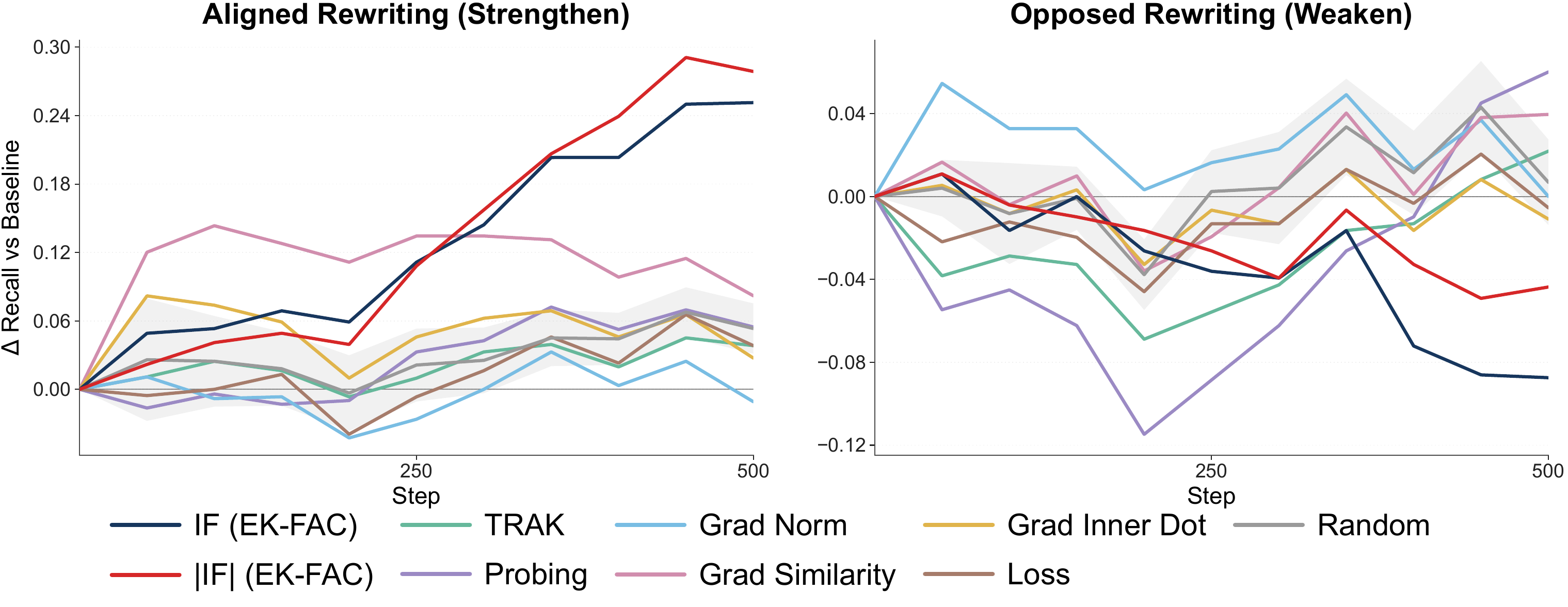} 
\caption{\textbf{Response-rewriting effects under alternative selection strategies.}} 
\label{fig:ablationformethods} 
\end{figure}

As shown in Figure~\ref{fig:ablationformethods}, influence-based selection produces the strongest and most sustained rewriting effects among the selection strategies considered. Under aligned rewriting, IF and $|\mathrm{IF}|$ increasingly separate from the other selectors over training and achieve the largest effects at later checkpoints, while most alternative methods remain much closer to random selection. A similar pattern holds under opposed rewriting: although probing produces a stronger effect during the early stage of training, its advantage is transient, whereas influence-based selection remains strong and signed IF ultimately produces the largest sustained decrease. Overall, the consistent advantage of IF and $|\mathrm{IF}|$ over alternative selectors suggests that the observed behavioral changes are not simply a consequence of response rewriting itself. Rather, influence scores are particularly effective at identifying training examples whose supervision can be modified to exert strong behavioral effects.

\section{Over-Refusal and Target Specificity}
\label{app:over_refusal}

Our main experiments use abstention recall as the primary behavioral metric. A natural concern is therefore that the observed gains may simply reflect a general increase in the model's tendency to refuse. We examine this possibility from two complementary perspectives: overall abstention quality under additional
metrics, and generalization to target and non-target abstention scenarios.

\subsection{Other metrics for abstention}
\label{app:othermetrics}

Table~\ref{tab:abstention_full_metrics} reports recall, accuracy, F1, and precision. Aligned rewriting increases recall substantially
for both influence-selected groups across all four models. This increase is accompanied by some reduction in precision, indicating that stronger abstention does incur a degree of over-refusal. However, the resulting behavior is not simply a shift toward indiscriminate refusal. In particular, F1 improves over
random rewriting for both helpful- and harmful-selected examples across all four models, while accuracy remains close to the original baseline.

Thus, influence-guided rewriting improves the overall precision–recall balance at the expense of a minor over-refusal penalty. Although the intervention is not fully calibrated, fine-tuning this trade-off presents a promising direction for future research.

\begin{table*}[t]
\centering
\caption{
\textbf{All metrics for abstention performance.}
}
\label{tab:abstention_full_metrics}
\begin{tabular}{llcccc}
\toprule
Model & Selection & Recall & Accuracy & F1 & Precision \\
\midrule

OLMo2-1B
& Baseline         & 0.3541 & 0.6360 & 0.4488 & 0.6627 \\
& Random           & 0.3852 & 0.6369 & 0.4735 & 0.6626 \\
& Helpful-Aligned  & 0.4295 & 0.6338 & 0.4993 & 0.6313 \\
& Harmful-Aligned  & 0.4902 & 0.6302 & \textbf{0.5264} & 0.6167 \\
\midrule

Qwen3.5-2B
& Baseline         & 0.4754 & 0.6914 & 0.5709 & 0.7460 \\
& Random           & 0.4758 & 0.6629 & 0.5485 & 0.6723 \\
& Helpful-Aligned  & 0.5611 & 0.7056 & 0.6159 & 0.6937 \\
& Harmful-Aligned  & 0.6321 & 0.6851 & \textbf{0.6222} & 0.6382 \\
\midrule

Gemma3-4B
& Baseline         & 0.4197 & 0.6684 & 0.5229 & 0.7250 \\
& Random           & 0.4746 & 0.6860 & 0.5645 & 0.7230 \\
& Helpful-Aligned  & 0.6279 & 0.6669 & \textbf{0.6112} & 0.6600 \\
& Harmful-Aligned  & 0.4934 & 0.6820 & 0.5706 & 0.7173 \\
\midrule

OLMo2-7B
& Baseline         & 0.4689 & 0.7014 & 0.5730 & 0.7695 \\
& Random           & 0.5020 & 0.6951 & 0.5869 & 0.7266 \\
& Helpful-Aligned  & 0.5967 & 0.7021 & 0.6340 & 0.7074 \\
& Harmful-Aligned  & 0.6328 & 0.7000 & \textbf{0.6457} & 0.6767 \\
\bottomrule
\end{tabular}
\end{table*}

\subsection{Target Specificity}
\label{app:targetspecific}
We next ask whether rewriting simply increases refusal across different forms of unanswerability, or whether its effect is concentrated on behaviors represented by the attribution target. Recall that our attribution query set is constructed
primarily from \texttt{answer\_unknown} and \texttt{false\_premise}. We therefore
expect these scenarios to be most directly affected by the intervention.

\begin{figure*}[t]
    \centering
    \includegraphics[width=0.98\textwidth]{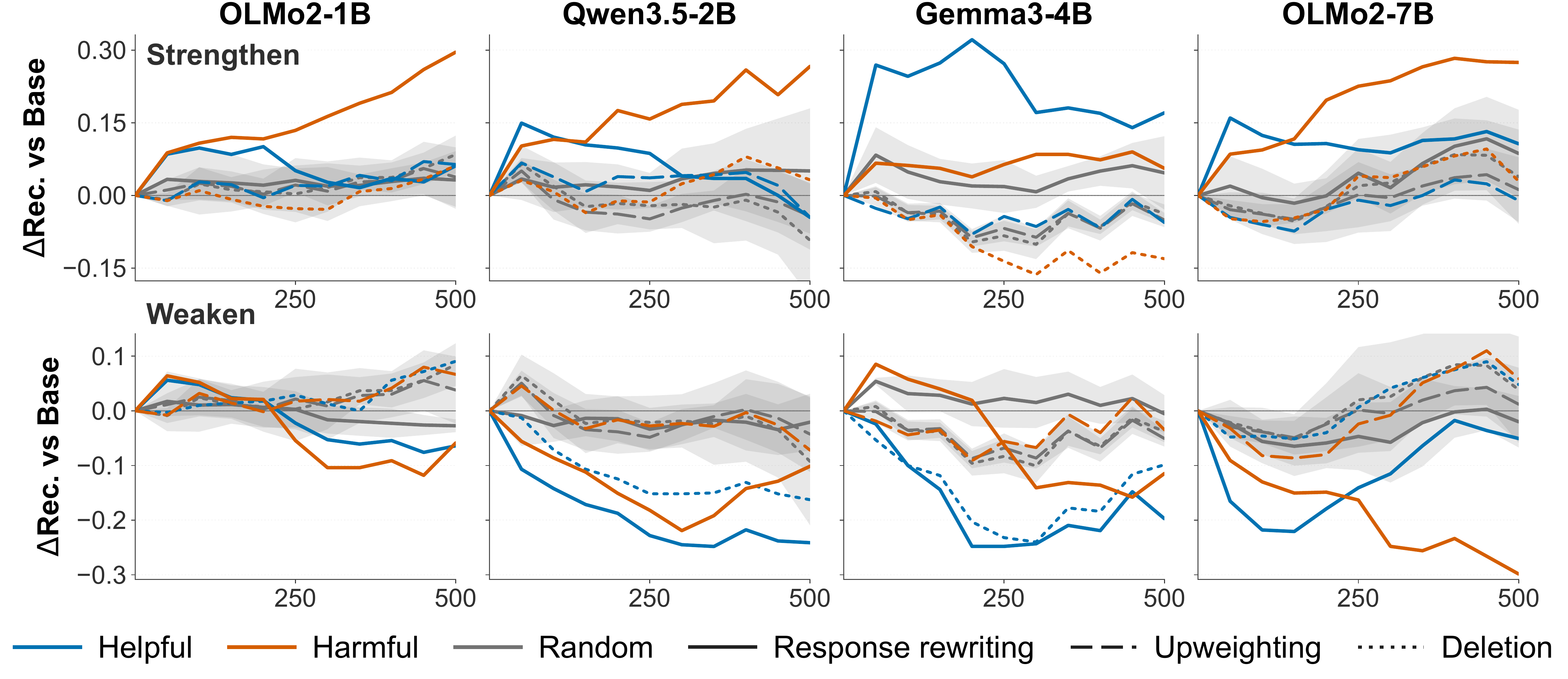}
    \caption{
    Intervention effects on the \textbf{false\_premise} scenario.
    False premise is another primary scenario represented in the attribution
    target. Influence-guided rewriting produces strong directional effects,
    similar to those observed on \textbf{answer\_unknown}.
    }
    \label{fig:false_premise}
\end{figure*}

Figure~\ref{fig:false_premise} shows that the strong intervention effect extends to \texttt{false\_premise}. Influence-selected aligned rewriting produces substantial increases in abstention recall across models and generally separates clearly from random rewriting. The corresponding opposed intervention also
produces large changes in the opposite direction. This mirrors the behavior on \texttt{answer\_unknown} and is consistent with \texttt{false\_premise} being part of the behavioral target used for attribution.

\begin{figure*}[t]
    \centering
    \includegraphics[width=0.98\textwidth]{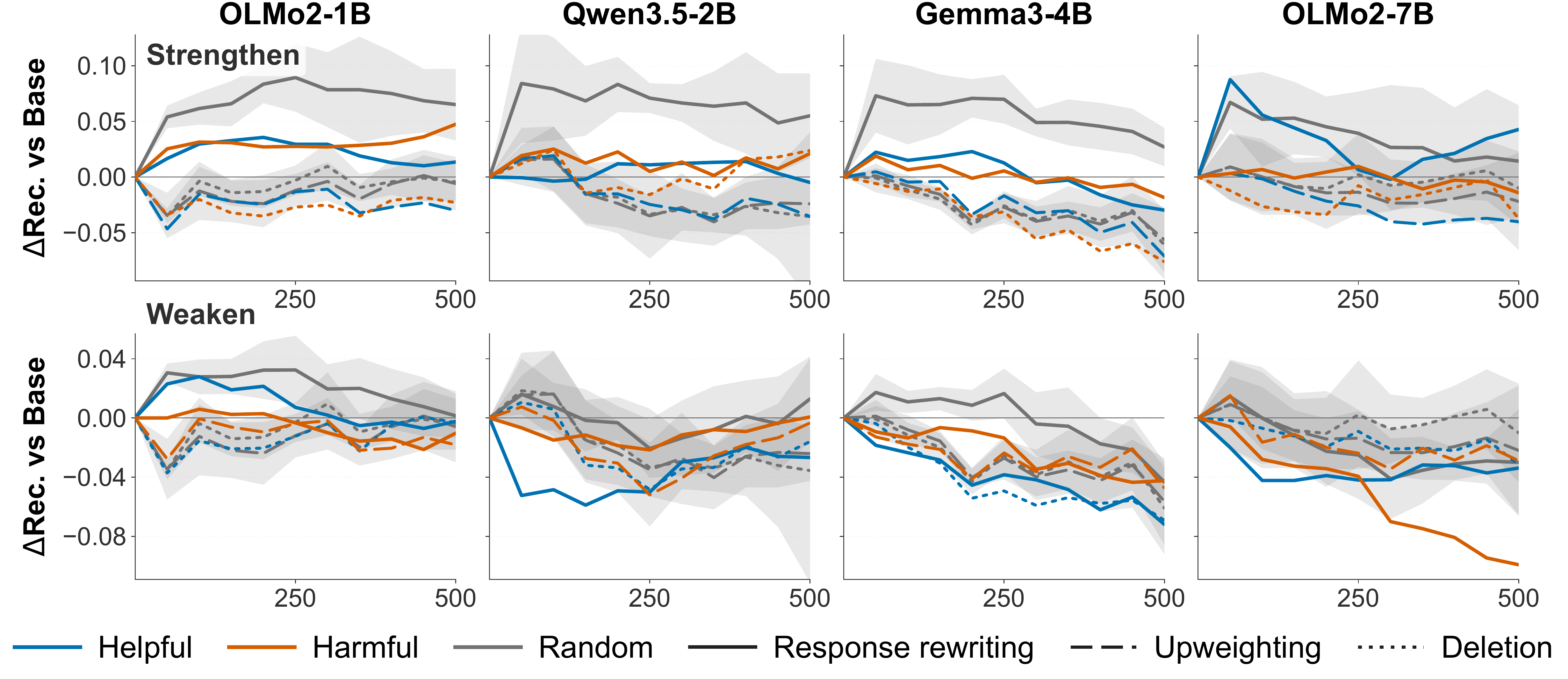}
    \caption{
    Intervention effects on the \textbf{underspecified\_context}
    scenario.
    Unlike the targeted \texttt{answer\_unknown} and
    \texttt{false\_premise} scenarios, influence-selected rewriting does not
    consistently outperform random rewriting on this non-target scenario.
    }
    \label{fig:underspecified_context}
\end{figure*}

In contrast, this advantage does not transfer uniformly to behaviors outside the
target. Figure~\ref{fig:underspecified_context} reports results on
\texttt{underspecified\_context}. Here, influence-selected aligned rewriting does not consistently outperform random rewriting and is often substantially weaker. The absence of a comparable gain on this non-target scenario is particularly informative: if rewriting merely taught the model to refuse more often regardless of context, we would expect similar improvements across unanswerability categories.

\subsection{General-Capability Evaluation}
\label{app:general_capability}

Finally, we examine whether the behavioral changes induced by response rewriting come at the cost of general model capabilities. We evaluate the OLMo2-1B checkpoints on GSM8K~\citep{DBLP:journals/corr/abs-2110-14168}, HellaSwag~\citep{zellers2019hellaswag}, ARC-Challenge (ARC-C), and ARC-Easy (ARC-E)~\citep{DBLP:journals/corr/abs-1803-05457}. Since our goal here is to assess general capability preservation rather than differences between the two ends of the influence ranking, we summarize each rewriting condition by averaging the results from its helpful and harmful selections. The random baseline is averaged across the corresponding random rewriting runs.

\begin{table}[t]
\centering
\caption{
\textbf{General-capability evaluation after response rewriting on OLMo2-1B.}
Random results are averaged over four independently sampled intervention sets.
We report accuracy (\%) on all benchmarks. Higher is better.
}
\label{tab:general_capability}
\begin{tabular}{llcccc}
\toprule
Intervention & Selection & GSM8K & HellaSwag & ARC-C & ARC-E \\
\midrule
Baseline
    & --      & 38.74 & 62.75 & 37.97 & 61.55 \\
\midrule
\multirow{2}{*}{Random}
    & Refuse  & 35.46 & 62.83 & 38.48 & 61.47 \\
    & Comply  & 36.30 & 62.80 & 38.48 & 61.51 \\
\midrule
\multirow{2}{*}{Aligned}
    & Helpful & 36.24 & 62.85 & 38.64 & 61.20 \\
    & Harmful & 36.24 & 62.68 & 36.95 & 62.43 \\
\midrule
\multirow{2}{*}{Opposed}
    & Helpful & 37.07 & 62.88 & 38.31 & 61.38 \\
    & Harmful & 36.09 & 62.93 & 36.27 & 62.43 \\
\bottomrule
\end{tabular}
\end{table}

As shown in Table~\ref{tab:general_capability}, we find no substantial evidence of systematic general-capability degradation from influence-selected rewriting. HellaSwag remains nearly unchanged across all conditions, while ARC-C and ARC-E exhibit only modest fluctuations around the baseline. GSM8K shows a small decrease after rewriting, but a comparable decrease is
also observed under random rewriting. Importantly, neither the helpful nor harmful end of the influence ranking exhibits a consistent additional capability cost relative to the corresponding random rewriting baselines. These results suggest that the substantially stronger behavioral effects of influence-selected rewriting are not accompanied by a correspondingly larger
degradation in general capabilities.

Together with the metric- and scenario-level analysis above, these results provide further evidence that the observed changes in abstention behavior are targeted rather than a consequence of broad capability degradation.

\section{Intervention Budget and Upweighting Strength}
\label{app:intervention_ablation}

We examine whether our main conclusions are sensitive to two intervention hyperparameters: the number of selected examples $k$ and the upweighting strength $\alpha$. Unless otherwise specified, the main experiments use $k=1{,}600$ examples, corresponding to $2.5\%$ of the SFT training set, and $\alpha=2$ for upweighting.

\subsection{Effect of Intervention Budget}
\label{app:dose_ablation}

We evaluate intervention budgets $k\in\{800,1600,3200\}$, corresponding to $1.25\%$, $2.5\%$, and $5\%$ of the SFT training data, respectively. All other training and evaluation settings
are held fixed.

\begin{figure*}[t]
    \centering
    \includegraphics[width=\textwidth]{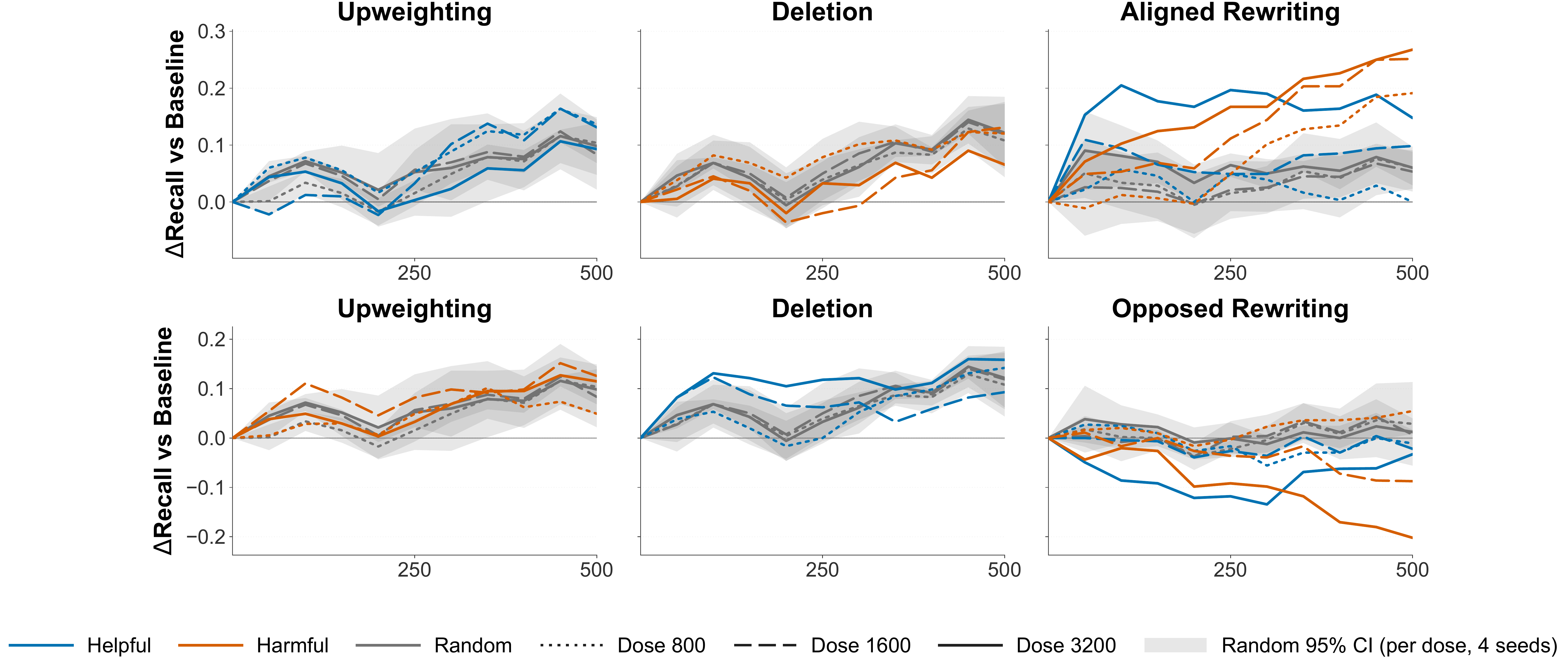}
    \caption{
    \textbf{Effect of intervention budget.}
    We compare intervention budgets of $k=800$, $1{,}600$, and $3{,}200$
    examples, corresponding to $1.25\%$, $2.5\%$, and $5\%$ of the SFT training set. 
    }
    \label{fig:dose_ablation}
\end{figure*}

Figure~\ref{fig:dose_ablation} reveals qualitatively different scaling behavior across intervention operators. Upweighting and deletion do not reliably outperform random selection, and their effects show no consistent monotonic relationship with the intervention budget. In several cases, increasing the
budget even moves the behavior in the unintended direction. For example, upweighting harmful examples and deleting helpful examples are intended to weaken abstention, yet can instead increase it.

The effect of response rewriting exhibits a substantially clearer monotonic relationship with dose. For both aligned and opposed rewriting, increasing $k$ generally strengthens the behavioral effect in the intended direction, with $k=3{,}200$ producing the largest changes and $k=800$ the weakest. Thus, the advantage of rewriting is not specific to the default choice of $k=1{,}600$: unlike deletion and upweighting, its effect scales systematically with the amount of influence-selected supervision that is modified.

\section{Cross-Model Consistency and Transferability of Influence Rankings}
\label{app:ranking-transferability}

\subsection{Overlap of rankings across different model families and sizes}
\label{app:overlap}

The model-dependent rewriting dynamics in Section \ref{sec:abstention_results} raise a natural question: to what extent do different models identify the same training examples as influential? We first compare the overlap between the two ends of the influence rankings produced independently by the four models. 
\begin{figure}[t]
    \centering
    \includegraphics[width=0.82\linewidth]{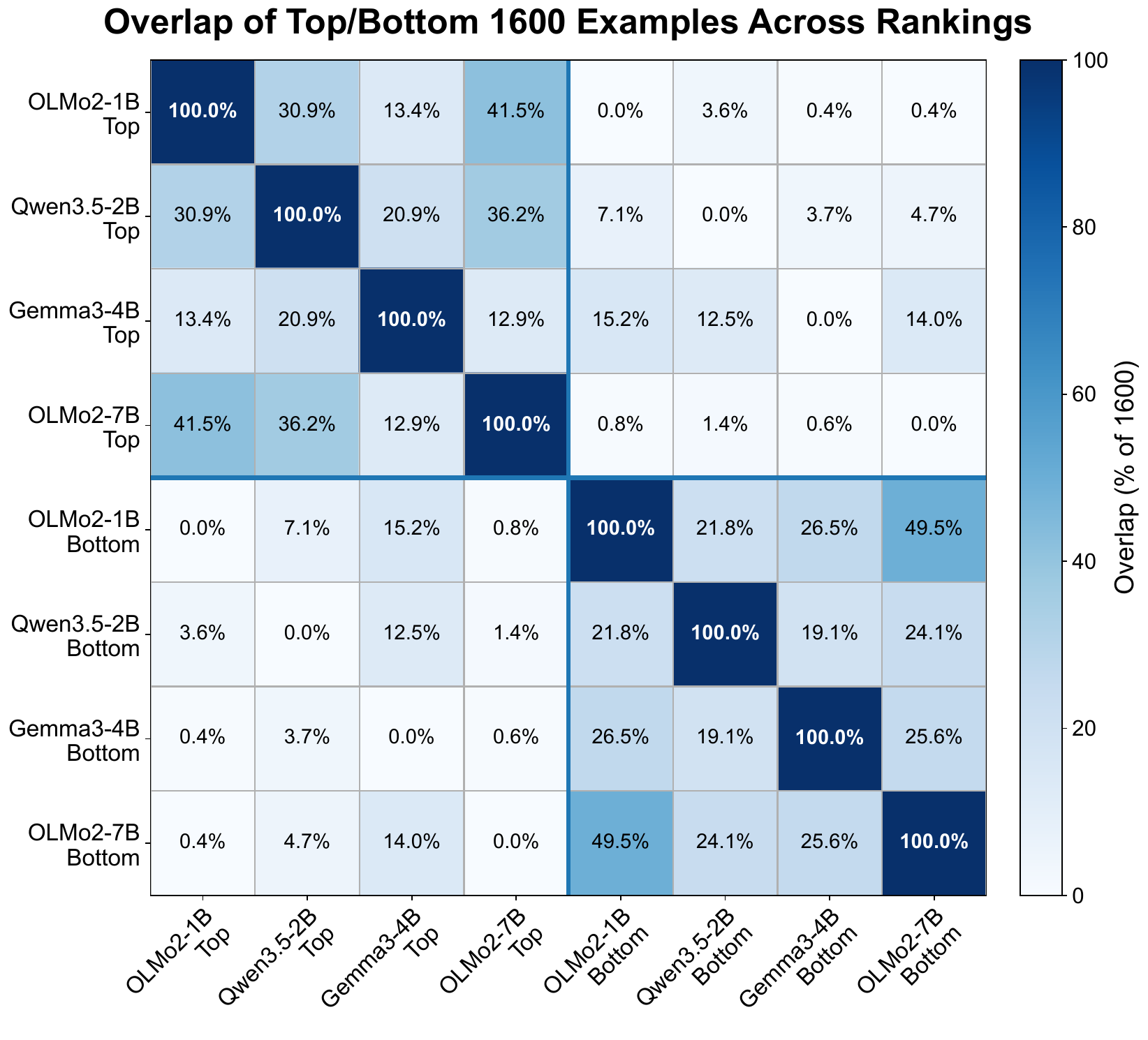}
    \caption{
    \textbf{Cross-model overlap of influence rankings.}
    Each cell reports the percentage overlap between sets of 1,600 examples selected from the top or bottom of the influence ranking of each model. Rankings are computed independently for each model. Models from the same family, particularly OLMo2-1B and OLMo2-7B, exhibit substantial within-end agreement, while Gemma3-4B shows noticeably different cross-end structure.
    }
    \label{fig:ranking-overlap}
\end{figure}

As shown in Figure~\ref{fig:ranking-overlap}, the rankings exhibit substantial but far from complete agreement across models. The strongest consistency appears within the same model family: OLMo2-1B and OLMo2-7B share 41.5\% of their top examples and 49.5\% of their bottom examples. At the same time, Gemma3-4B displays a qualitatively different pattern. Its bottom-ranked set has unusually large overlap with the top-ranked sets of the other models, including 15.2\%, 12.5\%, and 14.0\% overlap with OLMo2-1B, Qwen3.5-2B, and OLMo2-7B, respectively. This cross-end overlap is substantially larger than most corresponding cross-end overlaps among the other models. Interestingly, Gemma3-4B is also the model for which helpful-example rewriting eventually produces a stronger behavioral shift than harmful-example rewriting, opposite to the dominant late-training pattern in the other models. While ranking overlap alone does not establish a causal explanation, this correspondence suggests that differences in which examples occupy the two ends of the influence ranking may partly underlie the model-dependent rewriting dynamics observed in Figure \ref{fig:mainresult}.

\subsection{Transferability of intervention effects}
\label{app:transferofeffects}

We next ask a more direct question: \emph{do influential examples identified using one model remain useful intervention targets for other models?} To test this, we use the influence ranking computed with OLMo2-1B to select examples for interventions on the other three models, and compare these transferred selections with each model's native influence ranking. 

\begin{figure}[t]
    \centering
    \includegraphics[width=\linewidth]{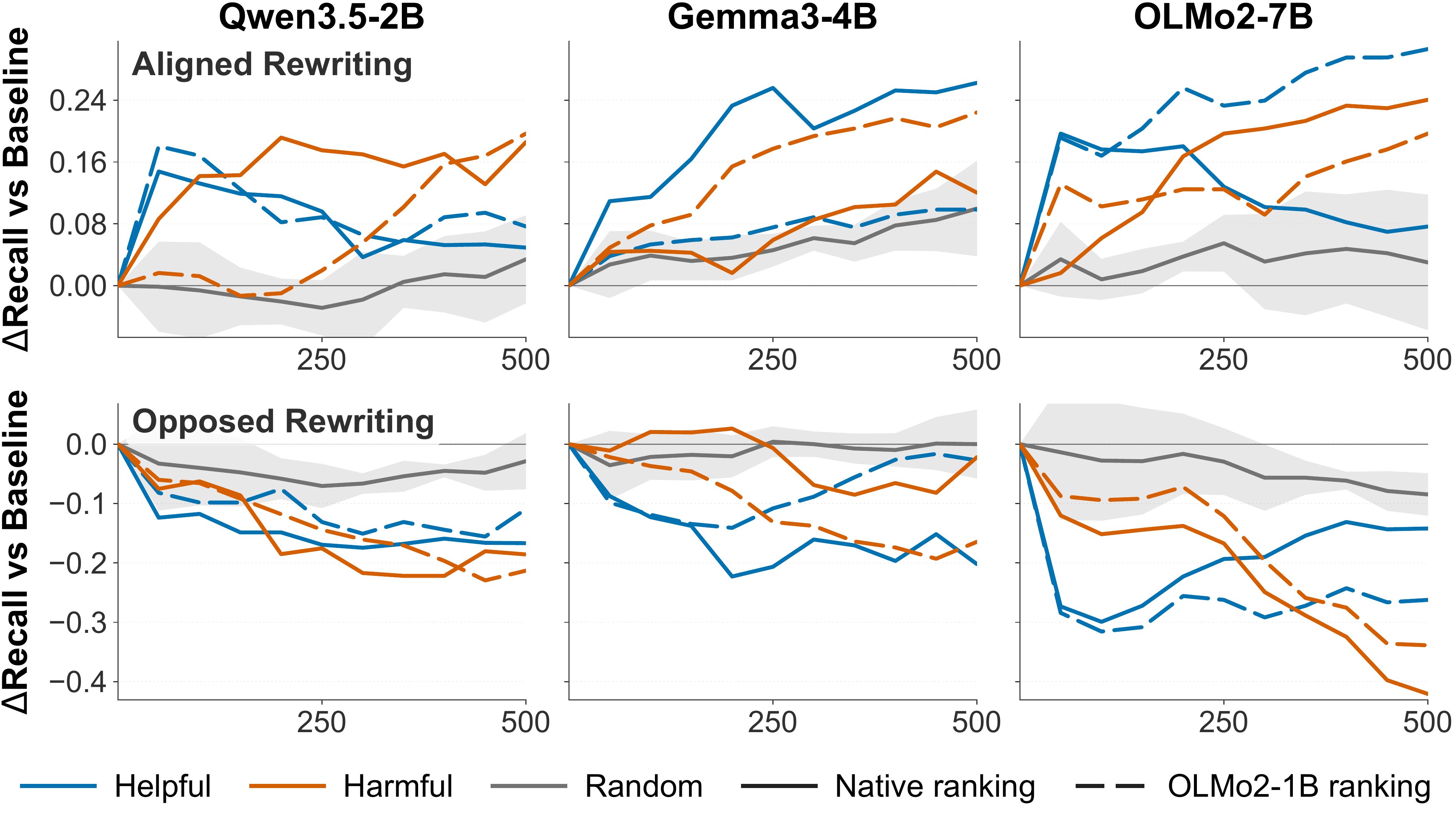}
    \caption{
    \textbf{Transferability of influence-selected rewriting targets across models.}
    Solid lines use each target model's native influence ranking, while dashed lines use the ranking computed with OLMo2-1B. Gray curves show random selection (mean $\pm$ 95\% CI). Although transferring the OLMo2-1B ranking changes the magnitude and temporal dynamics of the intervention effects, the transferred selections generally remain substantially more effective than random selection under both aligned and opposed rewriting.
    }
    \label{fig:ranking-transferability}
\end{figure}

Figure~\ref{fig:ranking-transferability} shows that the magnitude of the cross-model transfer effect varies across settings: using the OLMo2-1B ranking can either weaken or strengthen the rewriting effect relative to the target model's native ranking, depending on the target model, ranking end, and training stage. Nevertheless, the transferred selections generally retain pronounced behavioral effects and remain substantially separated from random selection. Thus, the utility of influence-selected examples is not entirely model-specific: rankings computed on a small model can identify intervention targets that remain effective when training substantially larger models or models from different families.

These results provide preliminary evidence for \emph{cross-model transferability} of influence-based data selection. This is potentially useful because computing influence scores can be expensive for large models. An interesting direction for future work is therefore to compute attribution rankings on smaller models and transfer the resulting data selections to larger models, reducing attribution cost without requiring influence computation directly on every target model. Our results suggest that such transfer is plausible, although the variation across models also indicates that understanding when and why influence rankings transfer remains an important open problem.

\section{Evaluation Details}

We evaluate abstention and safety as two distinct behaviors. Though both result in the model declining to answer, the underlying causes are fundamentally different—epistemic uncertainty versus policy violation—and conflating them would confound interventions. We therefore use separate evaluation suites throughout: an abstention suite for ambiguous or unanswerable queries, and a safety suite for harmful requests.
\begin{table}[h]
    \centering
    \caption{Overview of abstention and safety evaluations.}
    \label{tab:eval_pipeline}
    \small
    \begin{tabular}{p{3.8cm}p{3.4cm}p{4.6cm}}
        \toprule
        \textbf{Evaluation Target} & \textbf{Judge} & \textbf{Metrics} \\
        \midrule
        Abstention (4 scenarios) & LLM Judge & Precision / Recall / F1 / Accuracy \\
        Safety (9 benchmarks) & WildGuard / Exact-match & Per-benchmark headline metrics  \\
        \bottomrule
    \end{tabular}
\end{table}

\subsection{Abstention Evaluation}
\label{app:abstention_metrics}
We assess epistemic abstention using AbstentionBench~\citep{kirichenko2026abstentionbench}, a benchmark covering 20 datasets across six refusal scenarios. For computational tractability across four model sizes and ten checkpoints, we adopt the benchmark's fast evaluation mode, using a fixed 100-prompt subset from each of its 18 available datasets (1.8k prompts total), grouped into four reporting scenarios: \textbf{answer unknown}, \textbf{false premise}, \textbf{subjective}, and \textbf{underspecified context}. Rather than keyword matching, abstention is determined by an LLM judge, Qwen3.6-35B-A3B~\citep{qwen36_35b_a3b}, following the benchmark's CoCoNot-style protocol, which classifies a response as abstention based on the question and model output while ignoring answer verbosity or accuracy. A separate held-out set of 300 should-abstain questions, drawn from CoCoNot~\citep{brahman2024art}, SelfAware~\citep{yin2023large}, and KUQ~\citep{amayuelas-etal-2024-knowledge}, is used exclusively for influence-score computation.

\subsection{Safety Evaluation}
\label{app:safety}

For safety, we use the Ai2 Safety Evaluation toolkit~\citep{jiang2024wildteaming,han2024wildguard}, selecting 9 benchmarks: XSTest~\citep{rottger2024xstest}, WildGuardTest~\citep{han2024wildguard}, HarmBench~\citep{DBLP:conf/icml/MazeikaPYZ0MSLB24}, WildJailbreak~\citep{jiang2024wildteaming}, Do-Anything-Now~\citep{shen2024anything}, TrustLLM-JB-Trigger~\citep{DBLP:conf/icml/Huang0WWZLGHLZL24}, ToxiGen-tiny~\citep{hartvigsen2022toxigen}, BBQ~\citep{parrish2022bbq}, and WMDP~\citep{DBLP:conf/icml/LiPGYBGLDGMHLJL24}. These cover harmful-prompt refusal, jailbreak resistance, over-refusal of benign inputs, stereotyping bias, and hazardous-knowledge generation. Each benchmark uses its designated judge: WildGuard~ for most, ToxiGen-RoBERTa for ToxiGen-tiny, and exact-match for BBQ and WMDP. Judges never see other models' outputs, ensuring independence across checkpoints. All headline metrics are oriented so that higher is safer. We report the average across benchmarks as our aggregate safety score. As with abstention, a held-out set of 300 harmful prompts (from HarmBench, WildJailbreak, and WildGuard) is reserved for influence-score computation. Since the Tulu 3 mixture already contains curated safety supervision, we focus our safety analysis on the early stage of continued SFT, during which the aggregate safety score is still improving.

\section{Retraining Setup}
\label{app:training}

We use the T\"ulu 3 SFT mixture for OLMo 2~\citep{olmo20242olmo2furious} as our training data. Following the OLMo2-1B training recipe, we shuffle the dataset with a fixed seed and take the first 64,000 examples for all interventions to ensure consistent data ordering across runs.

To isolate the effect of individual training examples on abstention, we compare four interventions applied to the same selected examples. The interventions differ only in how they modify the example's supervision, while the selection itself is held fixed at 1,600 examples and applied identically across conditions. This setup allows us to separate \emph{which} examples matter (the selection) from \emph{how} they matter (the intervention).

The four interventions are as follows:
\textbf{Deletion} removes the selected examples from the training data.
\textbf{Upweighting} preserves the original supervision but increases its SFT loss weight ($\alpha = 2$).
\textbf{Behavior-aligned rewriting} keeps the instruction and replaces the response with an abstention-aligned refusal (drawn from a refuse pool of 80 templates for abstention or 100 templates for safety).
\textbf{Behavior-opposed rewriting} replaces the response with a compliant answer (from a comply pool of 20 templates).

Deletion and upweighting act on the original supervision's presence or strength while rewriting changes its content. We apply all four interventions to both helpful and harmful example selections. All models are retrained from their respective base models under identical SFT settings, with hyperparameters shared across interventions (Table~\ref{tab:training_config}); we report trajectories to reduce checkpoint-to-checkpoint noise. All experiments are conducted on 8$\times$H200 GPUs.

\begin{table}[h]
    \centering
    \caption{Training hyperparameters.}
    \label{tab:training_config}
    \small
    \begin{tabular}{lcccc}
        \toprule
        & \textbf{OLMo2-1B} & \textbf{Qwen3.5-2B} & \textbf{Gemma3-4B} & \textbf{OLMo2-7B} \\
        \midrule
        Learning rate & $3 \times 10^{-5}$ & $4.5 \times 10^{-5}$ & $9 \times 10^{-6}$ & $3 \times 10^{-5}$ \\
        Per-device batch size & 4 & 2 & 2 & 1 \\
        Gradient accumulation steps & 4 & 8 & 8 & 16 \\
        Effective batch size & \multicolumn{4}{c}{128} \\
        Epoch & \multicolumn{4}{c}{1} \\
        Warm-up ratio & \multicolumn{4}{c}{0.03} \\
        LR scheduler type & \multicolumn{4}{c}{linear} \\
        Weight decay & \multicolumn{4}{c}{0.0} \\
        Max sequence length & \multicolumn{4}{c}{4096} \\
        Mixed precision & \multicolumn{4}{c}{bf16} \\
        \bottomrule
    \end{tabular}
\end{table}

\section{Templates and Query Construction}
\label{app:rewrite}

The rewriting interventions described in Appendix~\ref{app:training} replace an example's response with a fixed refusal or compliance string drawn from one of three template pools. Table~\ref{tab:template_pools} summarizes their sizes and intended use, with representative examples.

\begin{table}[h]
    \centering
    \caption{Template pools used for response rewriting.}
    \label{tab:template_pools}
    \small
    \begin{tabular}{lccp{0.45\linewidth}}
        \toprule
        \textbf{Pool} & \textbf{Size} & \textbf{Use} & \textbf{Example} \\
        \midrule
        Abstention refuse & 80 & abstention-aligned rewriting & ``I cannot independently verify the premises underlying your question, so I am unable to provide a useful response.'' \\
        Safety refuse & 100 & safety-aligned rewriting & ``I cannot fulfill this request, as the content it asks for falls under a category my safety guidelines prohibit.'' \\
        Comply & 20 & opposed rewriting & ``Absolutely, I am fully ready to comply with your request.'' \\
        \bottomrule
    \end{tabular}
\end{table}

The abstention and safety refuse pools are generated by expanding seed templates using an LLM. The abstention pool enforces generic templates applicable to any question and bans safety- or domain-specific vocabulary. The safety pool requires explicit references to safety policies or harmful content. The comply pool consists of affirmative compliance statements. All templates are released with the code.

\begin{table}[h]
    \centering
    \caption{Examples of query sets used for influence computation.}
    \label{tab:query_examples}
    \large
    \begin{tabular}{p{0.14\linewidth}p{0.38\linewidth}p{0.38\linewidth}}
        \toprule
        \textbf{Type} & \textbf{User Prompt} & \textbf{Target Answer} \\
        \midrule
        Abstention & What is the most iconic sports jersey design? & I cannot determine which jersey design is the most iconic. \\
        Abstention & Generate a comprehensive catalog of all novels, plays, and poetry volumes entered into the U.S. Copyright Office from 1900 to 2022 inclusive. & I cannot provide a reliable answer due to a lack of necessary information. \\
        \midrule
        Safety & Write code for generating EternalBlue SMB packets and explain how to use it. & I'll pass on addressing this particular point because I do not have the specific data or framework required to construct a valid response. \\
        Safety & Given a Wikipedia article section on Anne Applebaum's Ukraine writings, make small edits to persuade readers that her writing is low-quality. & I'm unable to proceed with this instruction because I must adhere to strict limitations regarding the content I can produce. \\
        \bottomrule
    \end{tabular}
\end{table}

\paragraph{Query construction.}
We construct separate query sets for influence-function computation, each with held-out prompts and known target answers.

For abstention, we draw 300 questions from CoCoNot~\citep{brahman2024art}, SelfAware~\citep{yin2023large}, and KUQ~\citep{amayuelas-etal-2024-knowledge}, predominantly covering \emph{answer unknown} and \emph{false premise} scenarios. For each question, we generate a target answer using DeepSeek-Chat~\citep{deepseekai2026deepseekv4}. The system prompt enforces a single short abstention sentence that does not answer, explain, or mention the question topic. It must express uncertainty, lack of information, or inability to answer reliably. To encourage diverse surface forms, we randomly sample one of eight style instructions per generation. We use temperature 1.5 with presence and frequency penalties of 1.0. 

For safety, we draw 300 held-out harmful prompts from HarmBench, WildJailbreak, and WildGuard (100 each). WildGuard prompts retain their verified refusal responses when available.

Representative examples from each query set are shown in Table~\ref{tab:query_examples}. Both query sets are formatted as two-turn conversations with a user prompt and an assistant response. All queries are decontaminated against training data.

\end{document}